\documentclass[10pt,twocolumn,letterpaper]{article}

\usepackage[usenames,dvipsnames]{color}
\usepackage{cvpr}
\usepackage{times}
\usepackage{epsfig}
\usepackage{graphicx}
\usepackage{amsmath}
\usepackage{amssymb}

\usepackage{subfigure}
\usepackage{caption}
\usepackage{dsfont}
\usepackage{tabularx}
\usepackage{booktabs}
\usepackage{multirow}

\usepackage{soul}
\usepackage{color}
\newcommand{\comment}[1]{}

\newcommand{\note}[1]{\noindent\framebox{\begin{minipage}{8cm}{\bf #1}\end{minipage}}\\}

\newcommand{\bF}{\mathbf{F}}
\newcommand{\bH}{\mathbf{H}}
\newcommand{\bS}{\mathbf{S}}
\newcommand{\bW}{\mathbf{W}}
\newcommand{\bX}{\mathbf{X}}
\newcommand{\bh}{\mathbf{h}}

\newcommand{\bw}{\mathbf{w}}
\newcommand{\bx}[0]{\mathbf{x}}

\newcommand{\calL}{\mathcal{L}}
\newcommand{\calN}{\mathcal{N}}
\newcommand{\om}{\mathbf{\omega}}

\newcommand{\vect}[1]{#1}

\DeclareMathOperator*{\argmax}{arg\,max}

\newcommand{\fig}[1]{Fig.~\ref{fig:#1}}

\newcommand{\tbl}[1]{Table~\ref{tbl:#1}}

\definecolor{orange}{rgb}{1,0.5,0}
\definecolor{blue}{rgb}{0,0,0.6}

\newcommand{\kmyi}[1]{{\color{orange}{\bf #1}}}

\usepackage{snapshot}

\usepackage[pagebackref=true,breaklinks=true,letterpaper=true,colorlinks,bookmarks=false]{hyperref}

\cvprfinalcopy 

\def\cvprPaperID{1354} 

\ifcvprfinal\fi
\begin{document}

\title{TILDE: A Temporally Invariant Learned DEtector}

\author{Yannick Verdie\textsuperscript{1,}\thanks{First two authors contributed equally} \qquad  Kwang Moo Yi\textsuperscript{1,}\footnotemark[1]  \qquad  Pascal Fua\textsuperscript{1} \qquad    Vincent Lepetit\textsuperscript{2}\\
\textsuperscript{1}Computer Vision Laboratory, \'{E}cole Polytechnique F\'{e}d\'{e}rale de Lausanne (EPFL)\\
\textsuperscript{2}Institute for Computer Graphics and Vision, Graz University of Technology\\
{\tt\small \{yannick.verdie, kwang.yi, pascal.fua\}@epfl.ch, {\tt\small lepetit@icg.tugraz.at}}
}
\maketitle

\begin{abstract}

  
We introduce a  learning-based approach to detect  repeatable keypoints under
drastic  imaging   changes  of   weather  and   lighting  conditions   to  which
state-of-the-art  keypoint  detectors  are surprisingly  sensitive.   We  first
identify good  keypoint candidates  in multiple training  images taken  from the
same viewpoint.  We then  train a regressor to predict a  score map whose maxima
are those points so that they can be found by simple non-maximum suppression.


As there are no standard datasets to  test the influence of these kinds of changes,
we created our own, which we will make publicly available. We will show that our
method significantly outperforms the state-of-the-art methods in such challenging
conditions, while still achieving  state-of-the-art performance on untrained standard datasets.

\end{abstract}

\section{Introduction}

Keypoint detection  and matching is an  essential tool to address  many Computer
Vision  problems   such  as   image  retrieval,   object  tracking,   and  image
registration.  Since  the introduction  of the  Moravec, F\"orstner,  and Harris
corner detectors~\cite{Moravec80,Forstner87,Harris88} in  the 1980s, many others
have   been   proposed~\cite{Tuytelaars08,Forstner09,Rosten10}.   Some   exhibit
excellent repeatability  when the scale and  viewpoint change or the  images are
blurred~\cite{Mikolajczyk05}.  However, their reliability degrades significantly
when the images are acquired outdoors at different times of day and in different
weathers  or seasons,  as  shown  in Fig.~\ref{fig:teaser}.   This  is a  severe
handicap when attempting to match images taken  in fair and foul weather, in the
morning  and evening,  in winter  and summer,  even with  illumination invariant
descriptors~\cite{Gupta08b,Tang09,Gupta10,Wang11b}.

\def \teaserheight {9cm}
\begin{figure}
	\centering
	\subfigure[With SURF \cite{Bay08} keypoints]{
		 \includegraphics[height = \teaserheight , trim = 37 8 225 4, clip]{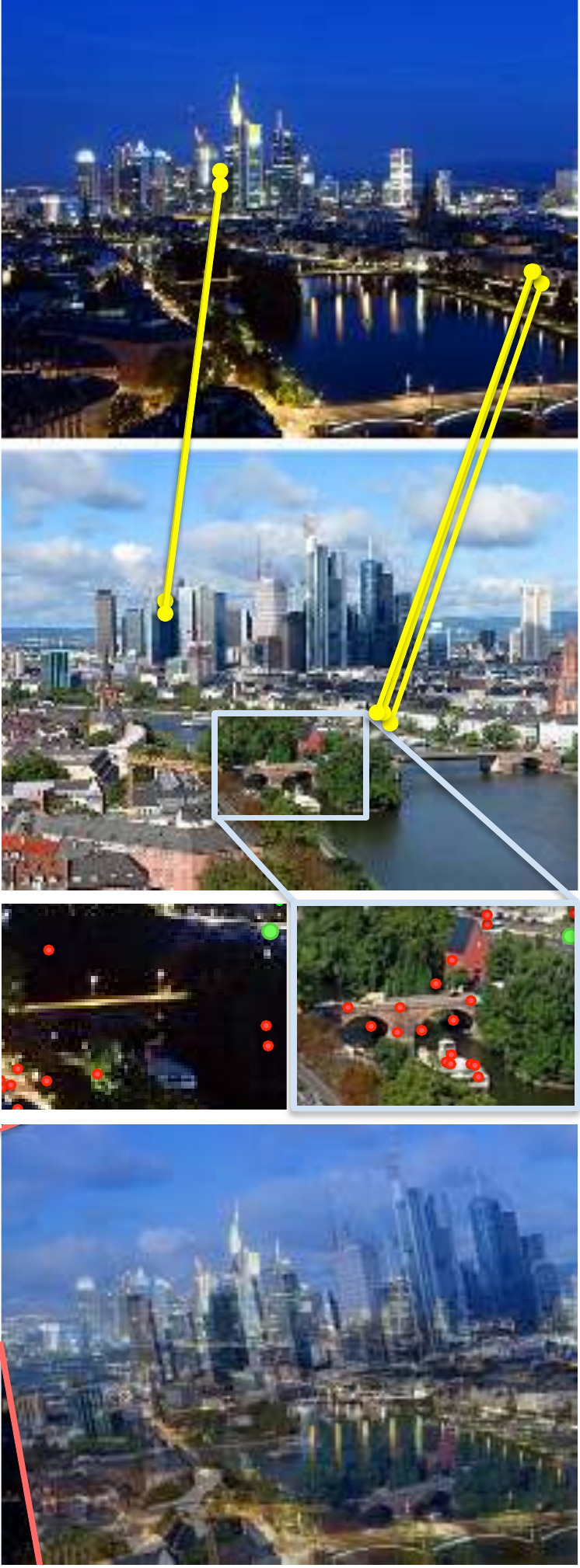}
	}
	\hspace{-0.4em} 
	\subfigure[With our keypoints]{
		\includegraphics[height = \teaserheight, trim = 255 8 4 4, clip]{figures/teaser/teaser_img.pdf}
	}
	
	\vspace{-0.5em}
	\caption{Image matching example using Speeded-Up Robust Features (SURF) \cite{Bay08} and our method. \emph{Same number of keypoints} and descriptor \cite{Lowe04} was used for both keypoint detectors. Detected keypoints are shown in the third row, with the repeated ones in green. For SURF, only one keypoint detected in the daytime image was
detected in the nighttime image. Our method on the other hand returns many common keypoints regardless of the drastic lighting change.
	\protect\footnotemark
	}
\label{fig:teaser}
	\vspace{-1.0em}
\end{figure}

\footnotetext{Figures are best viewed in color.}

In this paper, we propose an approach to learn a keypoint detector that extracts
keypoints which are stable under  such challenging conditions and allow matching
in situations as difficult as the one depicted by Fig.~\ref{fig:teaser}. To this
end, we  first introduce a simple  but effective method to  identify potentially
stable points in  training images.  We then  use them to train  a regressor that
produces  a score  map whose  values are  local maxima  at these  locations.  By
running  it on  new images,  we can  extract keypoints  with simple  non-maximum
suppression.    Our    approach   is    inspired   by   a    recently   proposed
algorithm~\cite{Sironi14} that relies on  regression to extract centerlines from
images of linear  structures.  Using this idea for our  purposes has required us
to  develop  a new  kind  of  regressor that  is  robust  to complex  appearance
variation so that it can efficiently and reliably process the input images.

As    in    the    successful     application    of    Machine    Learning    to
descriptors~\cite{Brown10,  Trzcinski13a}  and  edge  detection~\cite{Dollar06},
learning  methods  have  also  been  used before  in  the  context  of  keypoint
detection~\cite{Rosten06,Sochman07} to reduce the  number of operations required
when finding the {\it same} keypoints  as handcrafted methods. However, in spite
of    an   extensive    literature   search,    we   have    only   found    one
method~\cite{Strecha09} that attempts to improve the repeatability of keypoints by
learning.  This method focuses on learning  a classifier to filter out initially
detected keypoints but achieved limited  improvement.  This may be because their
method was  based on pure classification  and also because it  is non-trivial to
find good keypoints to be learned by a classifier in the first place.

Probably  as a  consequence, there  is currently  no standard  benchmark dataset
designed to test the robustness of keypoint detectors to these kinds of temporal
changes.  We  therefore created  our own  from images from  the Archive  of Many
Outdoor  Scenes~({\it AMOS})~\cite{Jacobs07}  and  our own  panoramic images  to
validate our approach.  We will use our dataset in addition to the standard {\it
  Oxford}~\cite{Mikolajczyk05}   and  {\it   EF}~\cite{Zitnick11}  datasets   to
demonstrate that our approach significantly outperforms state-of-the-art methods
in terms  of repeatability.  In  the hope of  spurring further research  on this
important topic, we will make it publicly available along with our code.

 In summary, our contribution is threefold:
 \begin{itemize}

 \item We introduce  a ``Temporally Invariant Learned DEtector''  (TILDE), a new
   regression-based approach  to extracting  feature points that  are repeatable
   under drastic illumination changes causes  by changes in weather, season, and
   time of day.

 \item We propose  an effective method to generate the  required training set of
   ``good keypoints to learn.''

 \item  We created  a  new benchmark  dataset for  evaluation  of feature  point
   detectors on outdoor images captured at different times ands seasons.

 \end{itemize}

 In the remainder of this paper, we first discuss related work, give an overview
 of our  approach, and  then detail our  regression-based approach.   We finally
 present the comparison of our approach to state-of-the-art keypoint detectors.


\section{Related Work}

\paragraph{Handcrafted Keypoint Detectors}

An extraordinary large amount of work has been dedicated to developing ever more
effective feature point detectors.  Even though the methods that appeared in the
1980s~\cite{Moravec80,Forstner87,Harris88} are still in  wide use, many new ones
have been developed since.  \cite{Forstner09}  proposed the SFOP detector to use
junctions as well  as blobs, based on a general  spiral model.  \cite{Hauagge12}
and  the  WADE detector  of~\cite{Salti13}  use  symmetries to  obtain  reliable
keypoints.   With  SIFER  and D-SIFER,  \cite{Mainali13,Mainali14}  used  Cosine
Modulated Gaussian filters and  10$^\text{th}$ order Gaussian derivative filters
for  more  robust  detection  of  keypoints.   Edge  Foci  \cite{Zitnick11}  and
\cite{Guan13} use edge information  for robustness against illumination changes.
Overall, these  methods have consistently  improved the performance  of keypoint
detectors on the standard  dataset~\cite{Mikolajczyk05}, but still suffer severe
performance drop when applied to outdoor scenes with temporal differences.

One of the major drawbacks of handcrafted methods are that they cannot be easily
adapted  to  the context,  and  consequently  lack flexibility.   For  instance,
SFOP~\cite{Forstner09}    works    well    when    calibrating    cameras    and
WADE~\cite{Salti13} shows good results when  applied to objects with symmetries.
However, their  advantages are not  easily carried on  to the problem  we tackle
here, such as finding similar outdoors scenes~\cite{Laffont14}.

\paragraph{Learned Keypoint Detectors}

Although work on keypoint detectors  were mainly focused on handcrafted methods,
some  learning   based  methods   have  already   been  proposed~\cite{Rosten06,
  Strecha09, Hartmann14,  Richardson13}.  With FAST,  \cite{Rosten06} introduced
Machine Learning techniques to learn  a fast corner detector.  However, learning
in their  case was  only aimed toward  the speed up  of the  keypoint extraction
process.    Repeatability   is  also   considered   in   the  extended   version
FAST-ER~\cite{Rosten10},   but   it   did   not   play   a   significant   role.
\cite{Strecha09}  trained  the  WaldBoost classifier~\cite{Sochman05}  to  learn
keypoints with high repeatability on a pre-aligned training set, and then filter
out an initial set of keypoints according to the score of the classifier.  Their
method, called  TaSK, is probably  the most related to  our method in  the sense
that  they use  pre-aligned  images to  build the  training  set.  However,  the
performance of their method is limited by the initial keypoint detector used.

Recently,  \cite{Hartmann14}  proposed  to  learn  a  classifier  which  detects
\emph{matchable} keypoints  for Structure-from-Motion~(SfM)  applications.  They
collect  \emph{matchable} keypoints  by observing  which keypoints  are retained
throughout the  SfM pipeline  and learn these  keypoints. Although  their method
shows significant  speed-up, they remain limited  by the quality of  the initial
keypoint  detector.  \cite{Richardson13}  learns  convolutional filters  through
random  sampling  and looking  for  the  filter  that  gives the  smallest  pose
estimation error when  applied to stereo visual  odometry.  Unfortunately, their
method  is  restricted  to  linear  filters,  which  are  limited  in  terms  of
flexibility, and it is not clear how  their method can be applied to other tasks
than stereo visual odometry.

We  propose a  generic  scheme  for learning  keypoint  detectors,  and a  novel
efficient  regressor   specified  for  this   task.   We  will  compare   it  to
state-of-the-art  handcrafted methods  as well  as TaSK,  as it  is the  closest
method from the literature, on several datasets.

\section{Learning a Robust Keypoint Detector}

\subfigcapmargin = 1.0em
\def \yop {3.15cm}
\begin{figure*}
\centering
	\subfigure[Stack of training images]{\label{fig:pillar}
	\includegraphics[height=\yop , trim = 7 2 12 10, clip]{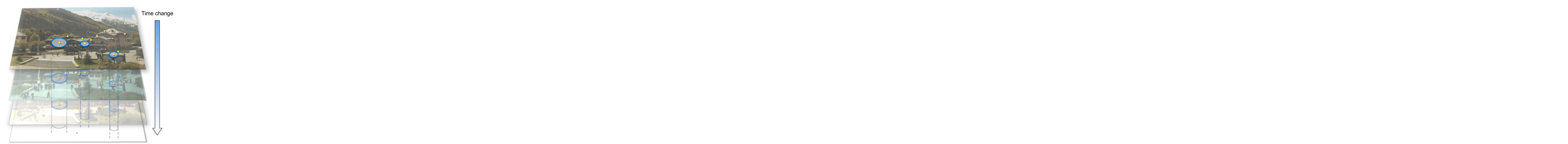}
	}	
 	\subfigure[Desired response on \mbox{positive samples}]{\label{fig:shape}
	\includegraphics[height=\yop , trim = 0.9cm 1.2cm 1.2cm 0.2cm, clip]{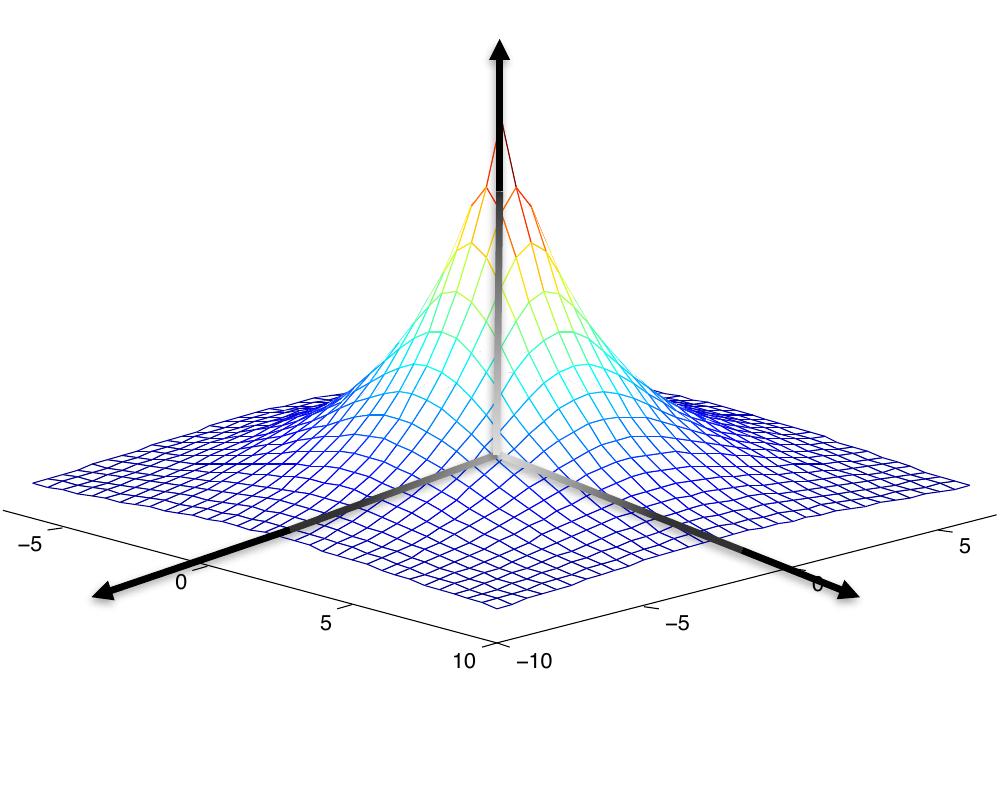}
	}
	\subfigure[Regressor response for a new image]{\label{fig:gt} 
	\includegraphics[height=\yop , trim = 5 5 8 5, clip]{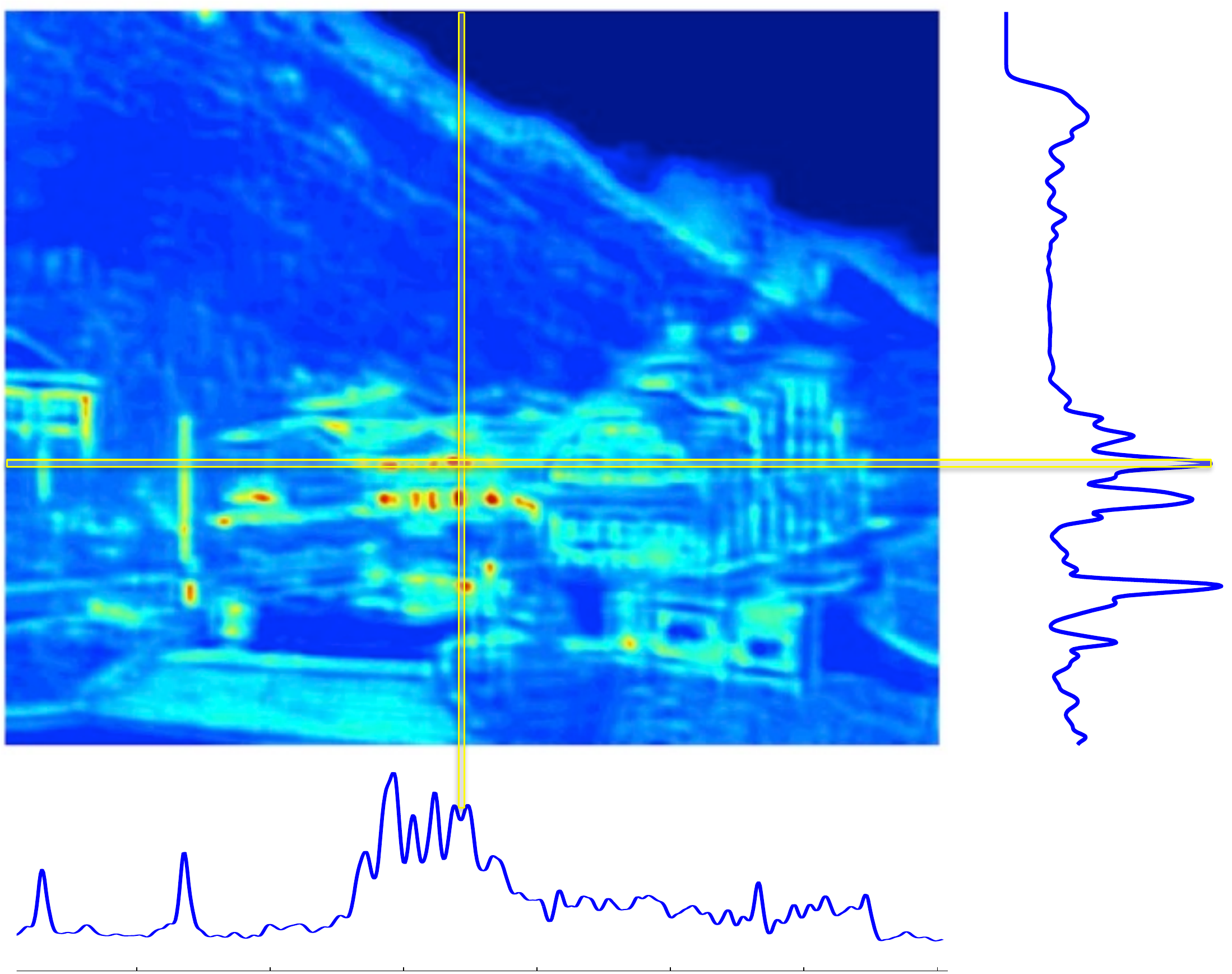}
	}
	\subfigure[Keypoints detected in the new image]{\label{fig:kp}
		\includegraphics[height=\yop, trim = 10 10 10 10, clip]{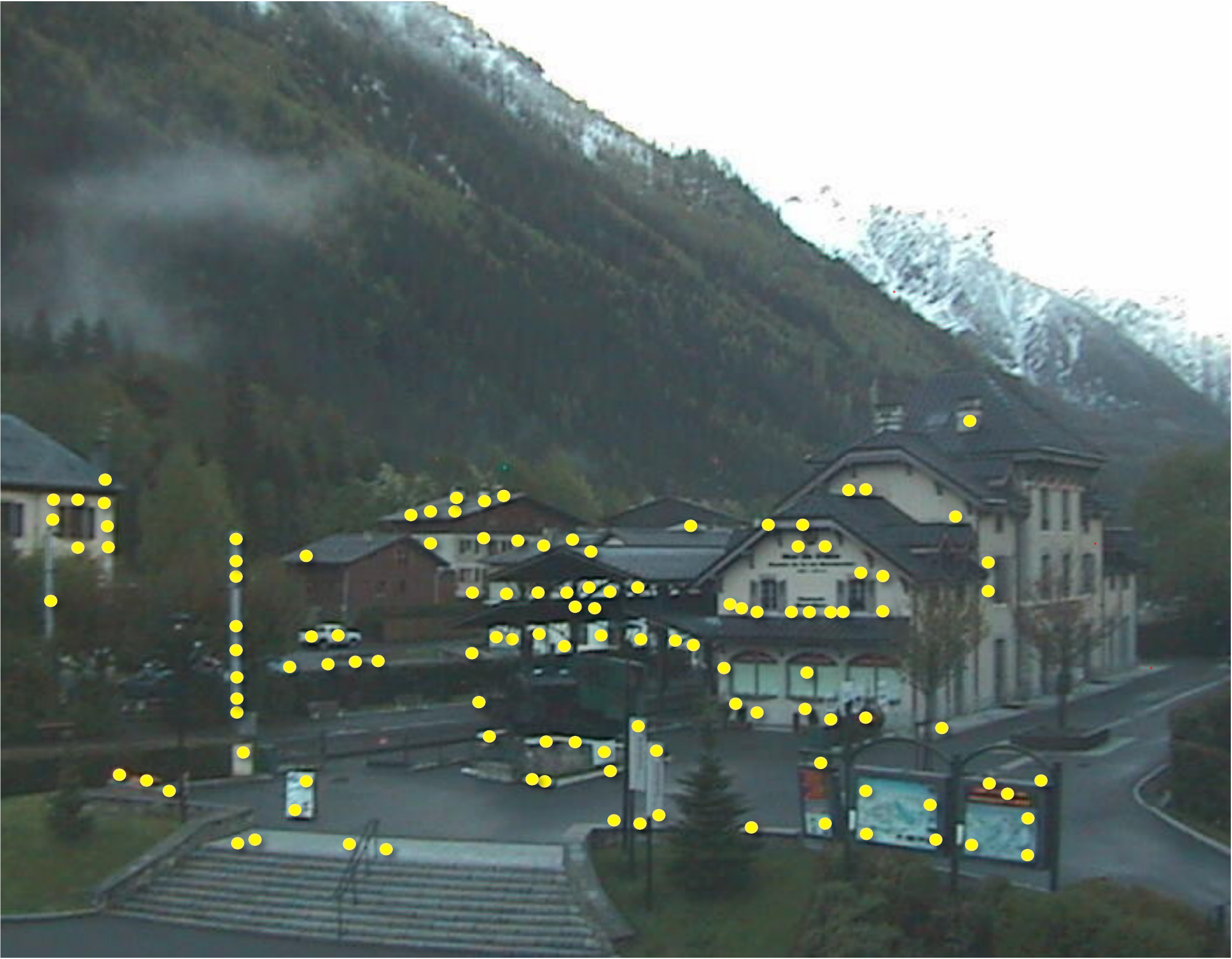}
	}
\vspace{-0.5em}
        \caption{Overview  of our  approach.  We  rely  on a  stack of  training
          images,  captured   from  the  same  viewpoint   but  under  different
          illuminations \subref{fig:pillar}, and a  simple method to select good
          keypoints to  learn. We train a  regressor on image patches  to return
          peaked values  like in  \subref{fig:shape} at the  keypoint locations,
          and small values far from  these locations. Applying this regressor to
          each  patch of  a new  image gives  us  a score  map such  as the  one
          in~\subref{fig:gt},  from  which  we   can  extract  keypoints  as  in
          \subref{fig:kp} by looking for local maxima with large values.}
\label{fig:pillar_illustration}

		\vspace{-0.5em}

\end{figure*}
\subfigcapmargin = .0cm

In this section, we first outline our regression-based approach briefly and then
explain how we build the required training set.  We will formalize our algorithm
and describe the regressor in more details in the following section.

\subsection{Overview of our Approach}

Let us  first assume that  we have a  set of training  images of the  same scene
captured from  the same  point of  view but at  different seasons  and different
times of the day,  such as the set of \fig{pillar}.  Let  us further assume that
we have identified in these images a set of locations that we think can be found
consistently over the  different imaging conditions. We propose  a practical way
of doing this in Section~\ref{sec:trainingset} below.
Let  us  call  \emph{positive  samples}  the image  patches  centered  at  these
locations in each training image.  The patches far away from these locations are
\emph{negative samples}.

To learn  to find these locations  in a new input  image, we propose to  train a
regressor to return a  value for each patch of a given size  of the input image.
These values should have a peaked shape  similar to the one shown in \fig{shape}
on the positive samples, and we also  encourage the regressor to produce a score
that is as small as possible for the negative samples.  As shown in \fig{gt}, we
can then  extract keypoints  by looking for local  maxima of  the values
returned by  the regressor, and discard  the image locations with  low values by
simple thresholding.  Moreover, our regressor  is also trained to return similar
values for the same locations over the  stack of images. This way, the regressor
returns consistent values even when the illumination conditions vary.





\def \datasetFigHeight {0.2}
\begin{figure*}
	\centering
	\subfigure[Sample images of the selected scenes from {\it AMOS}]{\label{fig:amosSix}
	\hspace{0.8em}
	 	\includegraphics[height=\datasetFigHeight\textwidth, trim = 10 10 0 0, clip]{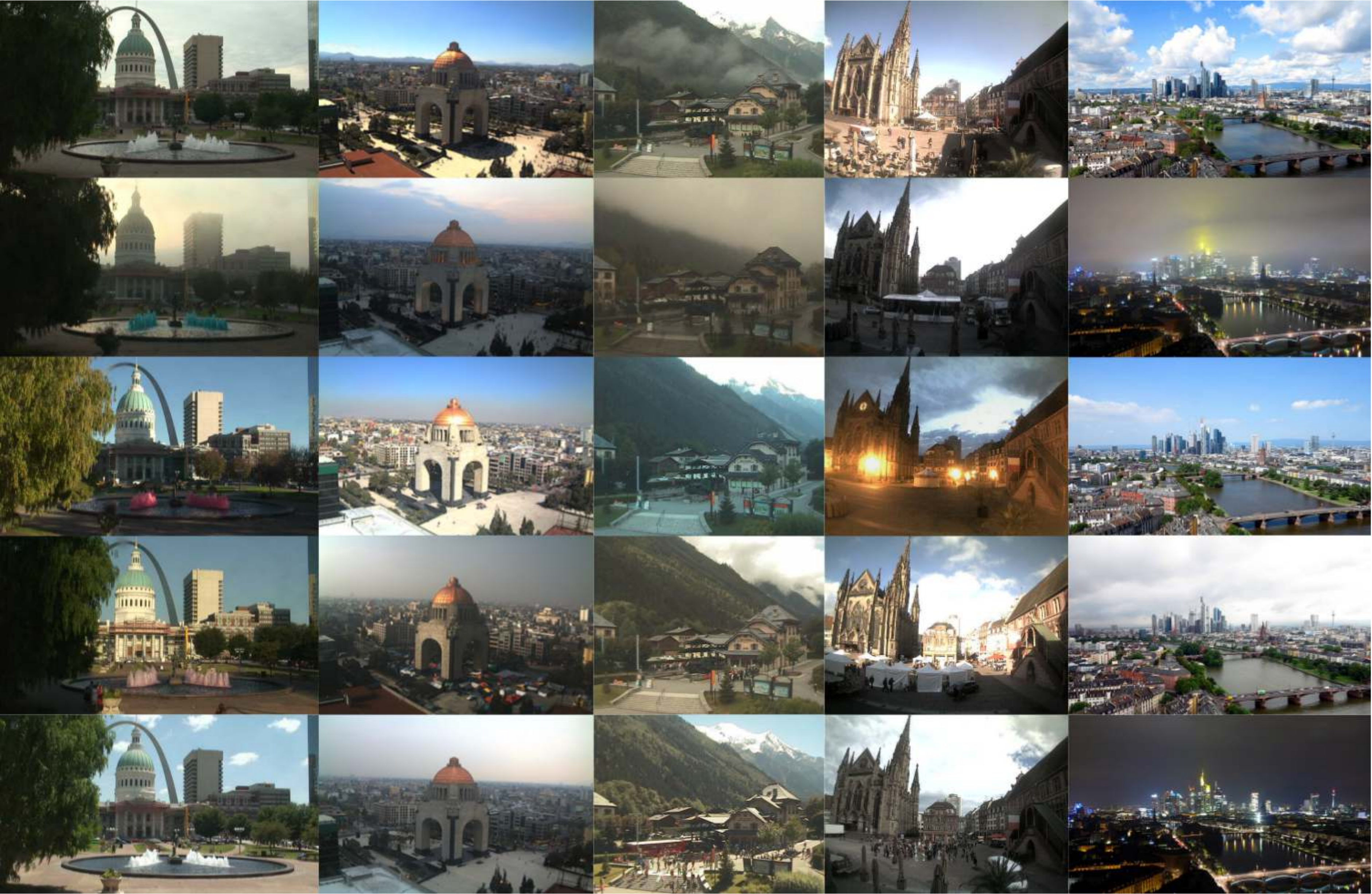}
		\hspace{0.8em}
	}
	\hspace{0.8em}
  	\subfigure[Sample       images      of       the      \emph{Panorama}
  	sequence]{\label{fig:panorama} \includegraphics[height=\datasetFigHeight\textwidth,
  	trim   =  10   10  10   10,  clip]{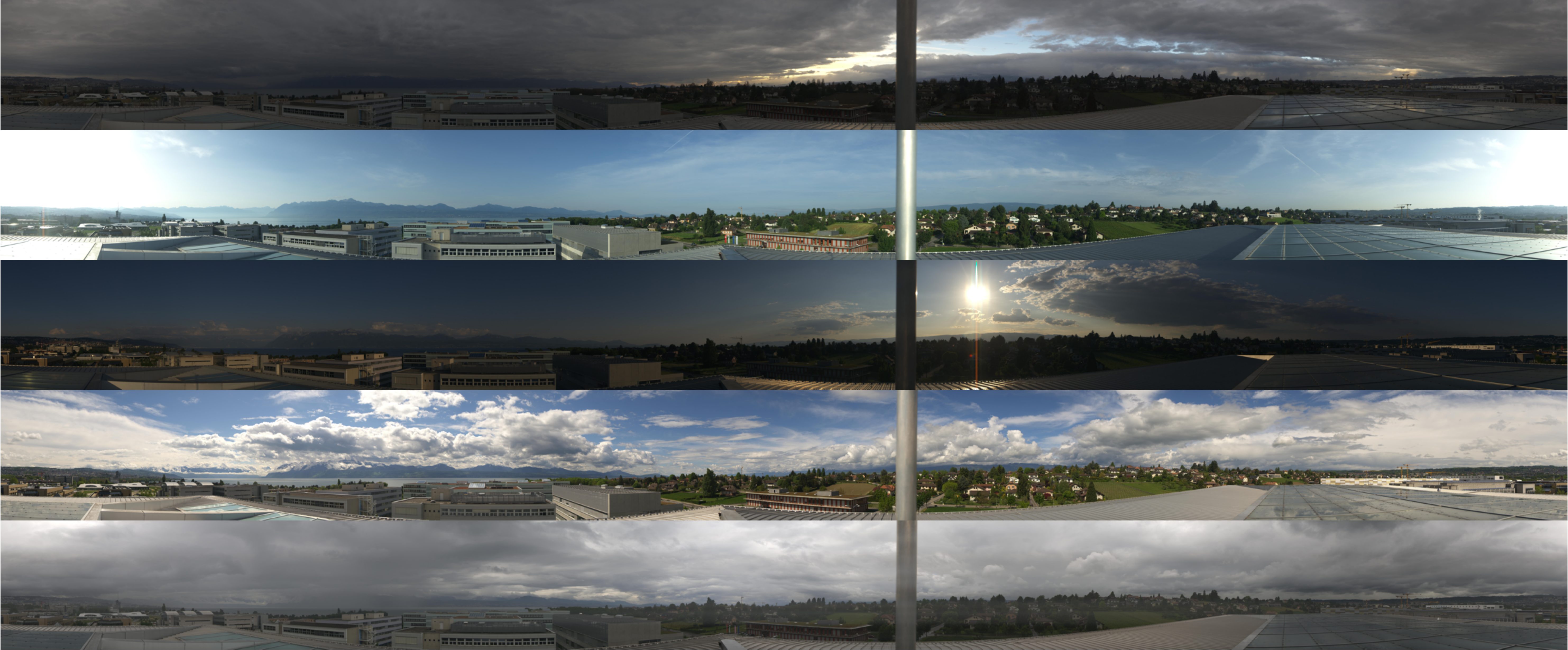}
  	}
	
    \vspace{-0.5em}  
     \caption{Example figures from  the \emph{Webcam} dataset. The {\it Webcam}
  	dataset  is  composed of  six  scenes from various locations: \subref{fig:amosSix}  five
  	scenes  taken   from  the   Archive  of   Many  Outdoor   Scenes~({\it AMOS})
  	dataset~\cite{Jacobs07},
  	namely \emph{StLouis}, \emph{Mexico}, \emph{Chamonix}, \emph{Courbevoie},
  	and \emph{Frankfurt}.  \subref{fig:panorama} {\it Panorama}  scenes from the roof of
  	a building which shows a 360 degrees view.}

\vspace{-0.9em}

	\label{fig:pillar_dataset}

\end{figure*}

\subsection{Creating the Training Set}
\label{subsec:GTKeypoints}
\label{sec:trainingset}

As shown in \fig{pillar_dataset}, to create our dataset of positive and negative
samples, we  first collected series of  images from outdoor webcams  captured at
different times  of day and  different seasons.  We identified  several suitable
webcams  from the  {\it  AMOS}  dataset~\cite{Jacobs07}---webcams that  remained
fixed over  long periods of  time, protected from the  rain, etc.  We  also used
panoramic images captured by a camera located on the top of a building.

To  collect a  training  set  of positive  samples,  we  first detect  keypoints
independently in  each image  of this dataset.   We use  SIFT~\cite{Lowe04}, but
other detectors could be considered as  well.  We then iterate over the detected
keypoints, starting with the keypoints with the smallest scale. If a keypoint is
detected at about the same location in  most of the images from the same webcam,
its location is likely to be a good candidate to learn.

In practice  we consider that  two keypoints are at  about the same  location if
their distance is smaller than the scale  estimated by SIFT and we keep the best
100 repeated locations. The set of positive  samples is then made of the patches
from  \emph{all} the  images,  including the  ones where  the  keypoint was  not
detected, and centered on the average location of the detections.


This simple strategy offers several advantages: we keep only the most repeatable
keypoints  for   training,  discarding   the  ones   that  were   detected  only
infrequently. We also  introduce as positive samples the patches  where a highly
repeatable keypoint  was missed.  This way,  we can focus on  the keypoints that
can be detected reliably under different conditions, and correct the mistakes of
the original detector.

To create  the set of negative  samples, we simply extract  patches at locations
that are far away from the keypoints used to create the set of positive samples.

\section{An Efficient Piece-wise Linear Regressor}

\label{sec:piecewiselinear}

In this section, we first introduce the  form of our regressor, which is made to
be  applied to  every patch  from  an image  efficiently, then  we describe  the
different terms of the proposed objective function to train for detecting keypoints reliably,  and finally we explain  how we optimize the  parameters of our
regressor to minimize this objective function.

\subsection{A Piece-wise Linear Regressor}

Our regressor is a piece-wise linear function expressed using Generalized
Hinging  Hyperplanes~(GHH)~\cite{Breiman93,Wang05d}:
\begin{equation}
\label{eq:piecewiseLinear}
\bF(\vect{\bx};  \vect{\om})  =   \sum_{n=1}^N  \delta_n  \max_{m=1}^M
\vect{\bw}_{nm}^\top\vect{\bx} \;\;,
\end{equation}
where $\vect{\bx}$  is a vector made  of image features extracted  from an image
patch, $\vect{\om}$  is the  vector of  parameters of the  regressor and  can be
decomposed into  $\vect{\om} = \left[\vect{\bw}^\top_{11},\ldots,\vect{\bw}^\top_{MN},
  \delta_1, \ldots,  \delta_N\right]^\top$.  The  $\vect{\bw}_{nm}$ vectors  can be
seen as linear  filters. The parameters $\delta_n$ are constrained  to be either
-1 or +1.  $N$  and $M$ are meta-parameters which control  the complexity of the
GHH.  As image features  we use the three components of the  LUV color space and
the  image  gradients---horizontal  and  vertical  gradients  and  the  gradient
magnitude---computed at each pixel of the $\vect{\bx}$ patches.

\cite{Wang05d}  showed  that any  continuous  piecewise-linear  function can  be
expressed in the  form of Eq.~\eqref{eq:piecewiseLinear}.  It is  well suited to
our keypoint  detector learning  problem, since applying  the regressor  to each
location of  the image  involves only simple  image convolutions  and pixel-wise
maximum operators, while regression trees require random access to the image and
the nodes,  and CNNs involve higher-order  convolutions for most of  the layers.
Moreover, we will show that this formulation also facilitates the integration of
different constraints, including constraints  between the responses for neighbor
locations,  which  are  useful  to  improve  the  performance  of  the  keypoint
extraction.

Instead of simply aiming to predict the  score computed from the distance to the
closest keypoint in a way similar  to what was done in~\cite{Sironi14}, we argue
that it is also  important to distinguish the image locations  that are close to
keypoints from  those that are far  away.  The values returned  by the regressor
for  image locations  close to  keypoints  should have  a local  maximum at  the
keypoint  locations, while  the actual  values for  the locations  far from  the
keypoints are  irrelevant as long  as they are small  enough to discard  them by
simple thresholding.   We therefore  first introduce a  classification-like term
that  enforces  the  separation  between  these two  different  types  of  image
locations.  We also  rely on a term  that enforces the response to  have a local
maximum at the keypoint locations, and  a term that regularizes the responses of
the regressor over time. To summarize, the objective function $\calL$ we minimize
over the parameters  $\om$ of our regressor  can be written as the  sum of three
terms:
\begin{equation}
\label{eq:problemFormulation}
\underset{\om}{\text{minimize}} \quad \calL_c(\om) + \calL_s(\om) + \calL_t(\om) \;\; .
\end{equation}


\subsection{Objective Function}

In  this subsection  we describe  in  detail the  three terms  of the  objective
function  introduced   in  Eq.~\eqref{eq:problemFormulation}.    The  individual
influences   of  each   term  are   evaluated  empirically   and  discussed   in
Section~\ref{sec:expThreeTerm}.

\paragraph{Classification-Like Loss $\calL_c$}

As explained  above, this term  is useful to  separate well the  image locations
that are  close to keypoints from  the ones that are  far away.  It relies  on a
max-margin loss, as in traditional  SVM~\cite{Cortes95}. In particular, we define
it as:
\begin{equation}
\label{eq:classificationLoss}
\calL_c(\om) = \gamma_c\left\|\vect{\om}\right\|_2^2
+
\frac{1}{K} \sum_{i=1}^{K}{ \max\left(0,1-y_i\bF\left(\vect{\bx}_i;\vect{\om}\right)\right)^2 } \;\; ,
\end{equation}
where $\gamma_c$  is a meta-parameter, $y_i  \in \left\{ -1, +1\right\}$  is the
label for  the sample $\bx_i$,  and $K$ is the  number of training
data.

\comment{   \kmyi{maybe   mention   recent   work  on   piecewise   linear   SVM
    \cite{Huang13}. Maybe we should mention it in the related works when talking
    about piecewise linear approaches.}  }

\paragraph{Shape Regularizer Loss $\calL_s$}

To have local maxima  at the keypoint locations, we enforce  the response of the
regressor to have a specific shape at these locations.  For each positive sample
$i$, we force the response shape by  defining a loss term related to the desired
response shape  $\bh$, similar to the  one used in~\cite{Sironi14} and  shown in
\fig{shape}:
\begin{equation}\label{eq:shapeformula}
\bh(x, y) = e^{\alpha (1-\frac{\sqrt{x^2+y^2}}{\beta})}-1 \;\; ,
\end{equation}
where $x$, $y$ are pixel coordinates with respect to the center of the
patch, and $\alpha$, $\beta$ meta-parameters  influencing the sharpness of the
shape.

However, we  want to enforce  only the  general shape and  not the scale  of the
responses  to not  interfere  with the  classification-like  term $\calL_c$.  We
therefore introduce an additional term defined as:
\begin{equation}
\label{eq:shapeRegularisor}
\calL_s(\om) =
\frac{\gamma_s}{K_p} \sum_{i|y_i=+1} \sum_{n} \left\| \bw_{n\eta_i(n)} * \bx_i - 
(\vect{\bw}_{n\eta_i(n)}^\top\vect{\bx_i})\bh\right\|_2^2 \;\; ,
\end{equation}
where  $*$ denotes  the convolution  product, $K_p$  is the  number of  positive
samples; $\gamma_s$  is a  meta-parameter for  weighting the  term that  will be
estimated          by          cross-validation.           $\eta_i(n)          =
\argmax_m\;\vect{\bw}_{nm}^\top\vect{\bx}_i$   is    used   to
enforce  the shape  constraints  only  on the  filters  that  contribute to  the
regressor response of the $\max$ operator.



It turns out that it is more convenient to perform the optimization of
this  term  in the  Fourier  domain.   If  we  denote the  2D  Fourier
transform of  $\bw_{nm}$, $\bx_i$,  and $\bh$ as  $\bW_{nm}$, $\bX_i$,
and $\bH$, respectively,  then by applying Parseval's  theorem and the
Convolution          theorem,          Eq.~\eqref{eq:shapeRegularisor}
becomes \footnote{ See  Appendix   in  the  supplemental   material  for
  derivation}\addtocounter{footnote}{-1}\addtocounter{Hfootnote}{-1}.
  
\begin{equation}
\label{eq:shapeRegularisorFourier}
\calL_s(\om) = \frac{\gamma_s}{K_p} \sum_{i|y_i=+1}\sum_n
\vect{\bW}^\top_{n\eta_i(n)}
\bS_i^\top \bS_i \vect{\bW_{n\eta_i(n)}} \;\; ,
\end{equation}
where
\begin{equation}
\label{eq:shapeRegularisorFourier_b}
\bS_i
= 
\left( \text{diag}\left(\vect{\bX_i}\right) - \vect{\bX}_i \vect{\bH} \right)^\top\;\; .
\end{equation}

\comment{
Through Parseval's  theorem and the  feature mapping proposed  in Ashraf~\etal's
work~\cite{Ashraf10},  we can easily combine  this  with the  other loss  terms
defined in the spatial domain\footnote{See Appendix in the supplemental material
  for details.}.
  }

This way  of enforcing  the shape of  the responses is  a generalization  of the
approach  of~\cite{Rodriguez13}  to   any  type  of  shape.    In  practice,  we
approximate  $\bS_i$  with the  mean  over  all  positive training  samples  for
efficient learning. We also use Parseval's  theorem and the feature mapping proposed in Ashraf~\etal's
work~\cite{Ashraf10} for easy calculation~\footnotemark .

\paragraph{Temporal Regularizer Loss $\calL_t$}

To enforce the repeatability of the  regressor over time, we force the regressor
to  have similar  responses at  the same  locations over  the stack  of training
images.  This is simply done by adding a term $\calL_t$ defined as:
\begin{equation}
\label{eq:manifoldRegularisor}
\calL_t(\om) =
\frac{\gamma_t}{K} \sum_{i=1}^{K} \sum_{j \in \calN_i}
\left(
\bF(\vect{\bx}_i;\vect{\om})-
\bF(\vect{\bx}_j;\vect{\om})\right)^2 \;\; ,
\end{equation}
where $\calN_i$ is the set of samples at the same image locations as $\bx_i$ but
from  the   other  training  images  of   the  stack.  $\gamma_t$  is   again  a
meta-parameter to weight this term.

\subsection{Learning the Piece-wise Linear Regressor}


\def \separableFigWidth {0.098}
\def \separableFigyopWidth{0.06} 
\def \nameData {StLouis}

\begin{figure}
	\centering
	\subfigure{
		  	\includegraphics[width=\separableFigWidth \textwidth, trim = 25 650 32 58, clip]{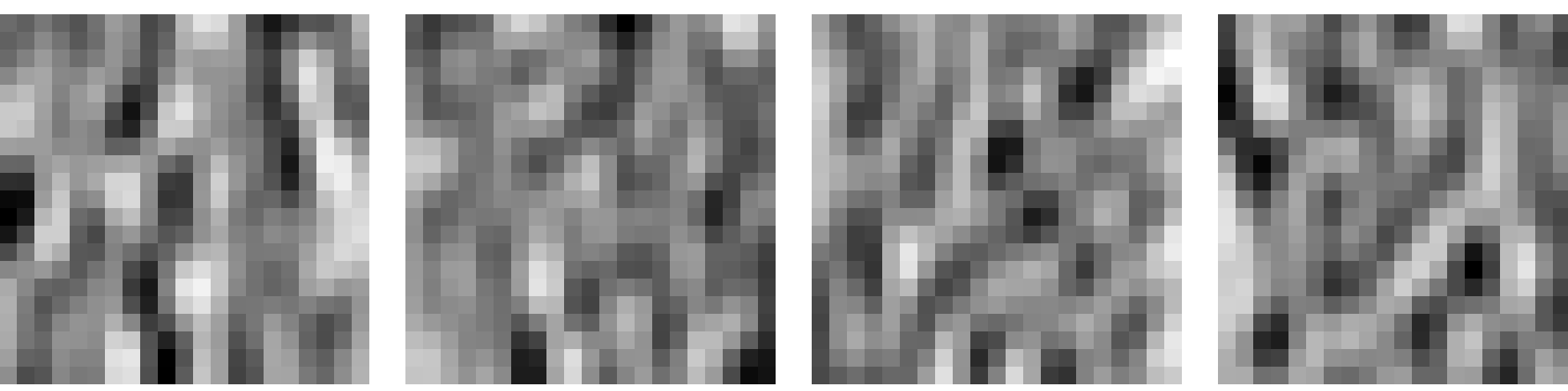} \hspace{-0.3em}
		  	\includegraphics[width=\separableFigWidth \textwidth, trim = 25 510 32 200, clip]{figures/separable/\nameData/original_rank004_dim1.pdf} \hspace{-0.3em}
		  	\includegraphics[width=\separableFigWidth \textwidth, trim = 25 369 32 340, clip]{figures/separable/\nameData/original_rank004_dim1.pdf} \hspace{-0.3em}
		  	\includegraphics[width=\separableFigWidth \textwidth, trim = 25 230 32 480, clip]{figures/separable/\nameData/original_rank004_dim1.pdf} \hspace{-0.3em}
	}

	\vspace{-0.9em}
	\subfigure{
		  	\includegraphics[width=\separableFigWidth \textwidth, trim = 25 650 32 60, clip]{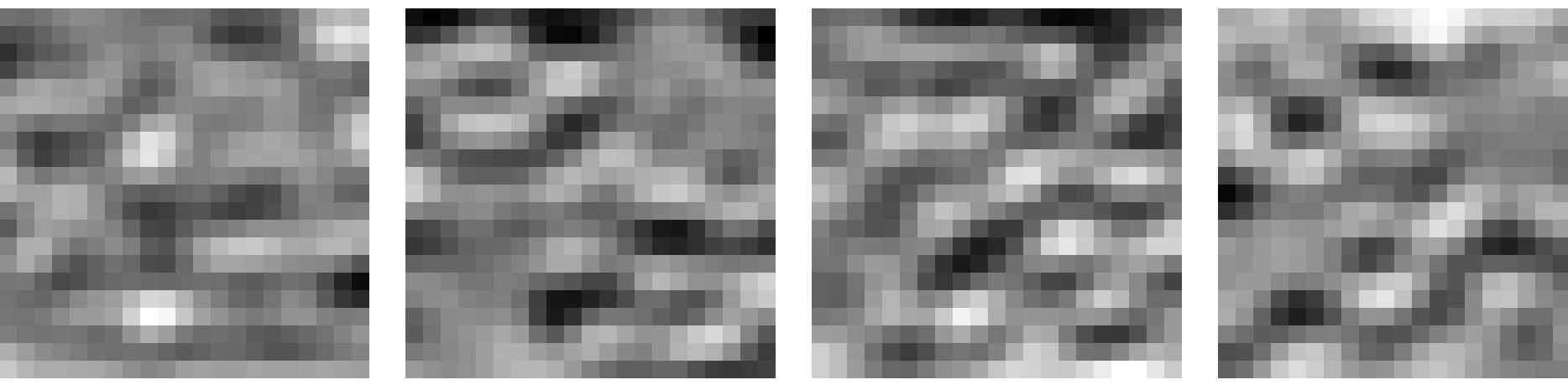} \hspace{-0.3em}
		  	\includegraphics[width=\separableFigWidth \textwidth, trim = 25 510 32 200, clip]{figures/separable/\nameData/original_rank004_dim2.pdf} \hspace{-0.3em}
		  	\includegraphics[width=\separableFigWidth \textwidth, trim = 25 369 32 340, clip]{figures/separable/\nameData/original_rank004_dim2.pdf} \hspace{-0.3em}
		  	\includegraphics[width=\separableFigWidth \textwidth, trim = 25 230 32 480, clip]{figures/separable/\nameData/original_rank004_dim2.pdf} \hspace{-0.3em}
	}

	\vspace{-0.9em}
	\subfigure{
		  	\includegraphics[width=\separableFigWidth \textwidth, trim = 25 650 32 60, clip]{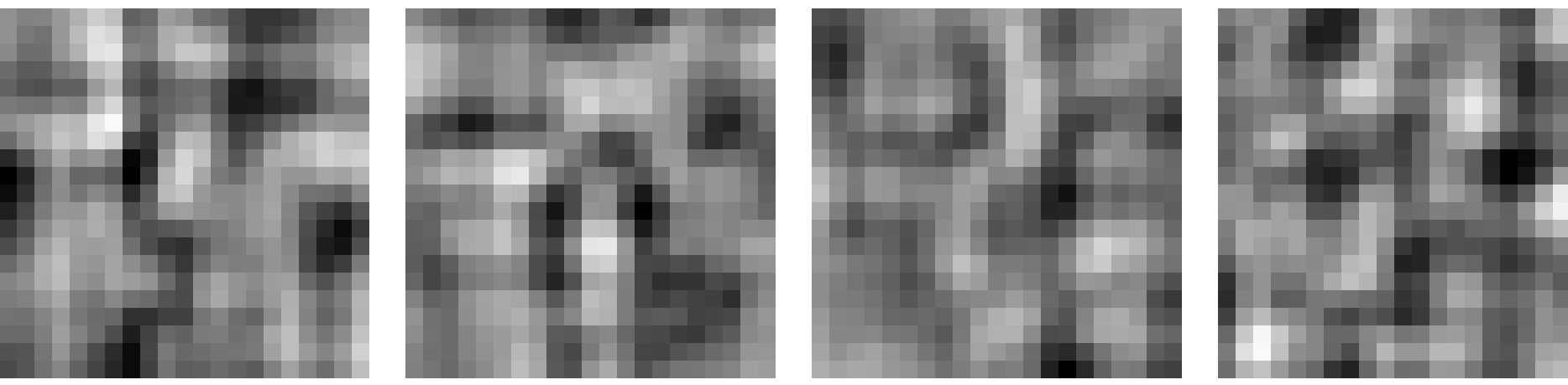} \hspace{-0.3em}
		  	\includegraphics[width=\separableFigWidth \textwidth, trim = 25 510 32 200, clip]{figures/separable/\nameData/original_rank004_dim3.pdf} \hspace{-0.3em}
		  	\includegraphics[width=\separableFigWidth \textwidth, trim = 25 369 32 340, clip]{figures/separable/\nameData/original_rank004_dim3.pdf} \hspace{-0.3em}
		  	\includegraphics[width=\separableFigWidth \textwidth, trim = 25 230 32 480, clip]{figures/separable/\nameData/original_rank004_dim3.pdf} \hspace{-0.3em}
	}

	\vspace{-0.9em}
	\subfigure{
		  	\includegraphics[width=\separableFigWidth \textwidth, trim = 25 650 32 60, clip]{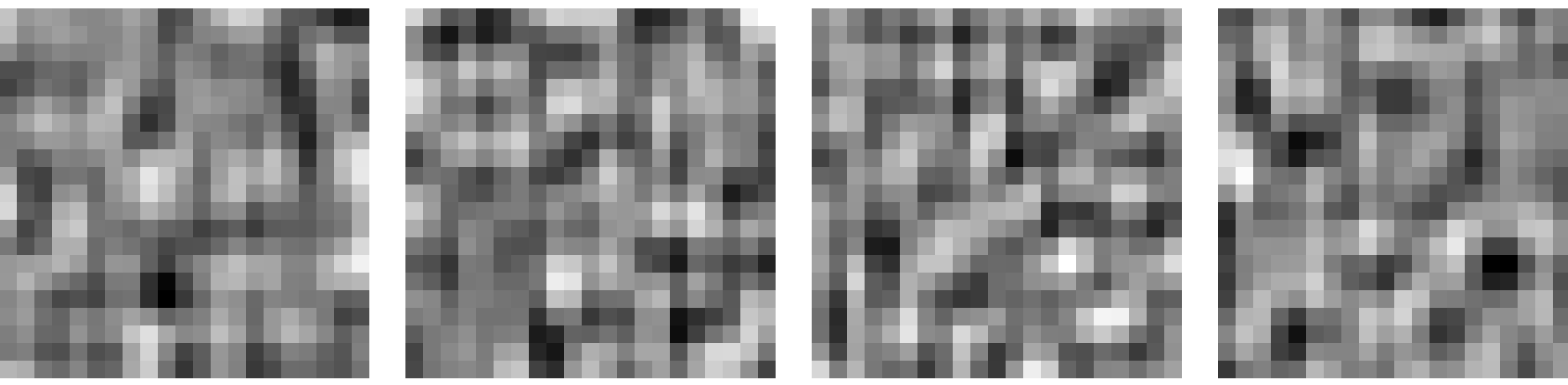} \hspace{-0.3em}
		  	\includegraphics[width=\separableFigWidth \textwidth, trim = 25 510 32 200, clip]{figures/separable/\nameData/original_rank004_dim4.pdf} \hspace{-0.3em}
		  	\includegraphics[width=\separableFigWidth \textwidth, trim = 25 369 32 340, clip]{figures/separable/\nameData/original_rank004_dim4.pdf} \hspace{-0.3em}
		  	\includegraphics[width=\separableFigWidth \textwidth, trim =25 230 32 480, clip]{figures/separable/\nameData/original_rank004_dim4.pdf} \hspace{-0.3em}
	}

	\vspace{-0.9em}
	\subfigure{
		  	\includegraphics[width=\separableFigWidth \textwidth, trim = 25 650 32 60, clip]{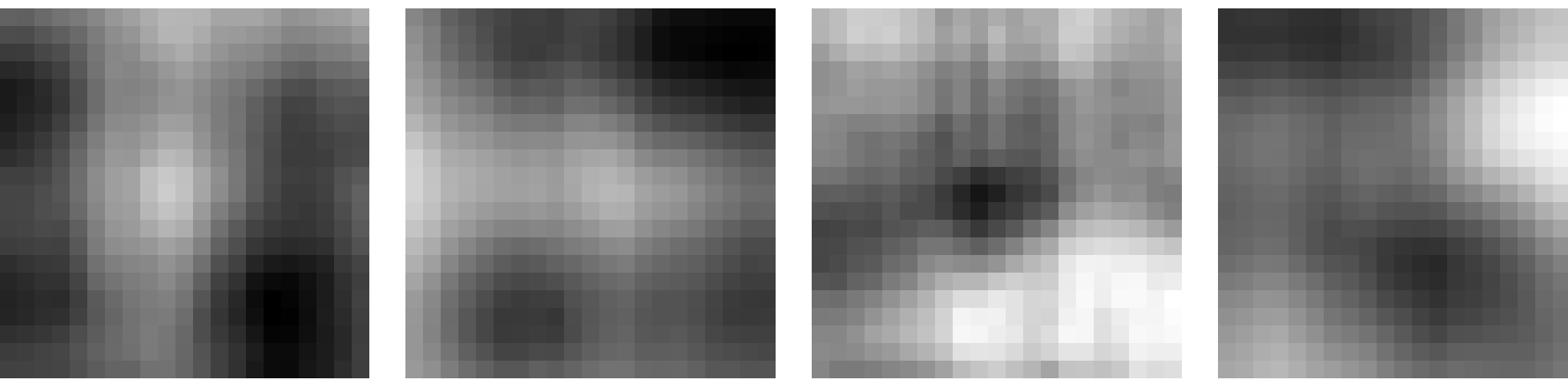} \hspace{-0.3em}
		  	\includegraphics[width=\separableFigWidth \textwidth, trim = 25 510 32 200, clip]{figures/separable/\nameData/original_rank004_dim5.pdf} \hspace{-0.3em}
		  	\includegraphics[width=\separableFigWidth \textwidth, trim = 25 369 32 340, clip]{figures/separable/\nameData/original_rank004_dim5.pdf} \hspace{-0.3em}
		  	\includegraphics[width=\separableFigWidth \textwidth, trim = 25 230 32 480, clip]{figures/separable/\nameData/original_rank004_dim5.pdf} \hspace{-0.3em}
	}

	\vspace{-0.9em}
	\setcounter{subfigure}{0}
	\subfigure[Original filters]{
		  	\includegraphics[width=\separableFigWidth \textwidth, trim = 25 650 32 60, clip]{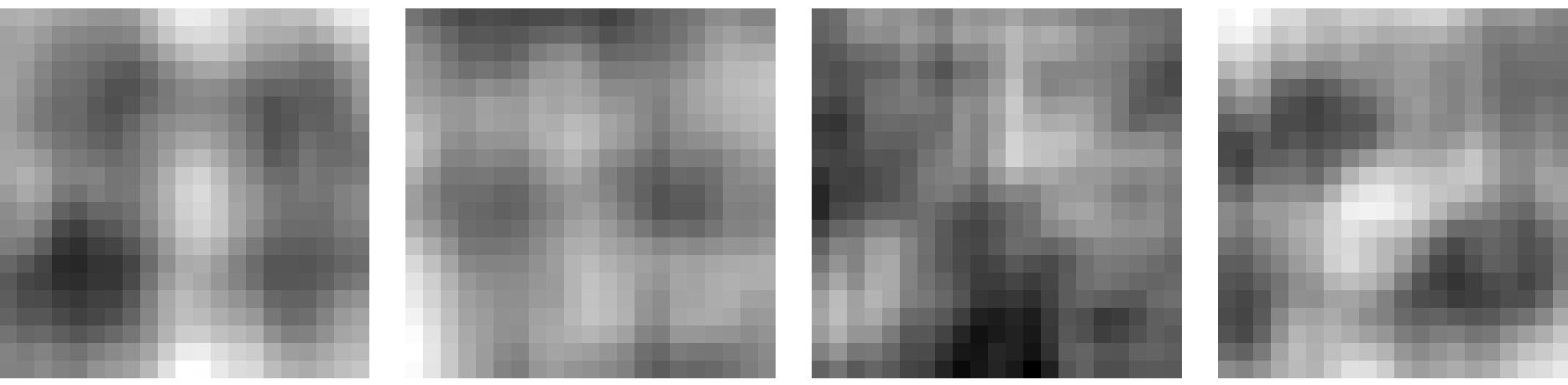} \hspace{-0.3em}
		  	\includegraphics[width=\separableFigWidth \textwidth, trim = 25 510 32 200, clip]{figures/separable/\nameData/original_rank004_dim6.pdf} \hspace{-0.3em}
		  	\includegraphics[width=\separableFigWidth \textwidth, trim = 25 369 32 340, clip]{figures/separable/\nameData/original_rank004_dim6.pdf} \hspace{-0.3em}
		  	\includegraphics[width=\separableFigWidth \textwidth, trim = 25 230 32 480, clip]{figures/separable/\nameData/original_rank004_dim6.pdf} \hspace{-0.3em}
	\label{fig:separableOrig}	
	}


	\subfigure{
		  	\includegraphics[width=\separableFigyopWidth \textwidth, trim = 25 510 25 65, clip]{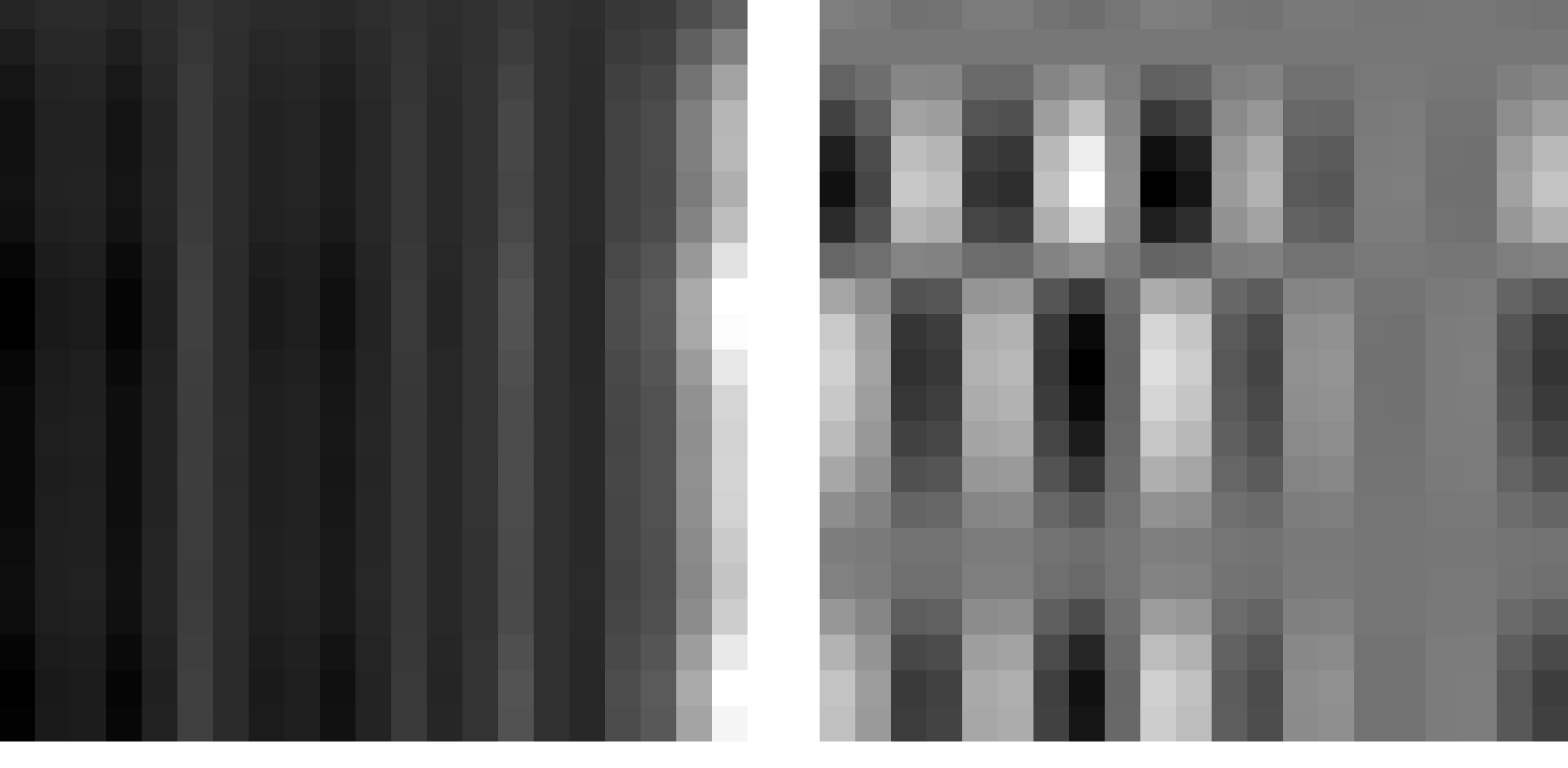} \hspace{-0.3em}
		  	\includegraphics[width=\separableFigyopWidth \textwidth, trim = 25 225 25 330, clip]{figures/separable/\nameData/separable_rank004_dim1.pdf} \hspace{-0.3em}

			\hspace{0.3em}
	
		  	\includegraphics[width=\separableFigyopWidth \textwidth, trim = 25 510 25 65, clip]{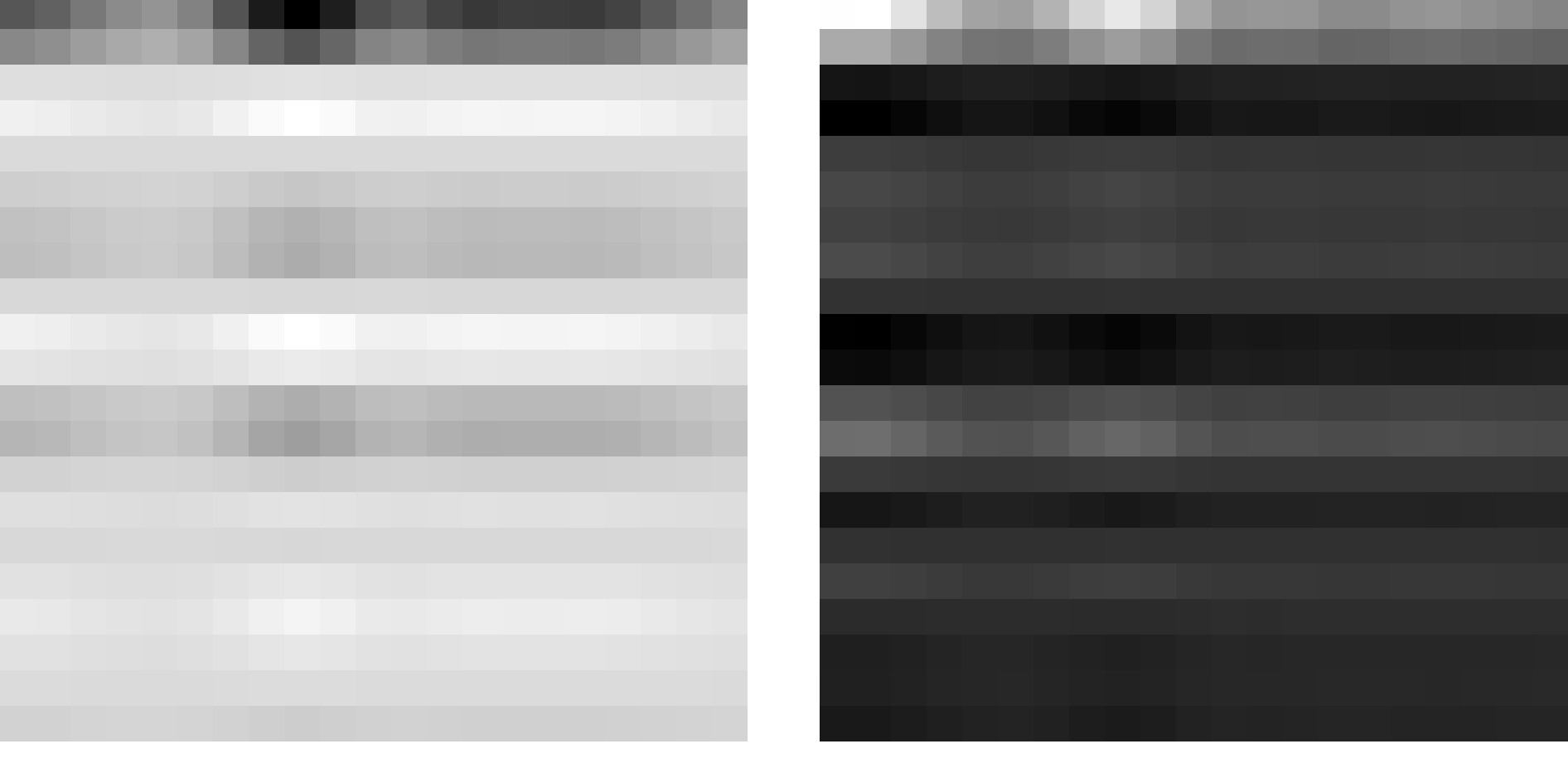} \hspace{-0.3em}
		  	\includegraphics[width=\separableFigyopWidth \textwidth, trim = 25 225 25 330, clip]{figures/separable/\nameData/separable_rank004_dim2.pdf} \hspace{-0.3em}

			\hspace{0.3em}

		  	\includegraphics[width=\separableFigyopWidth \textwidth, trim = 25 510 25 65, clip]{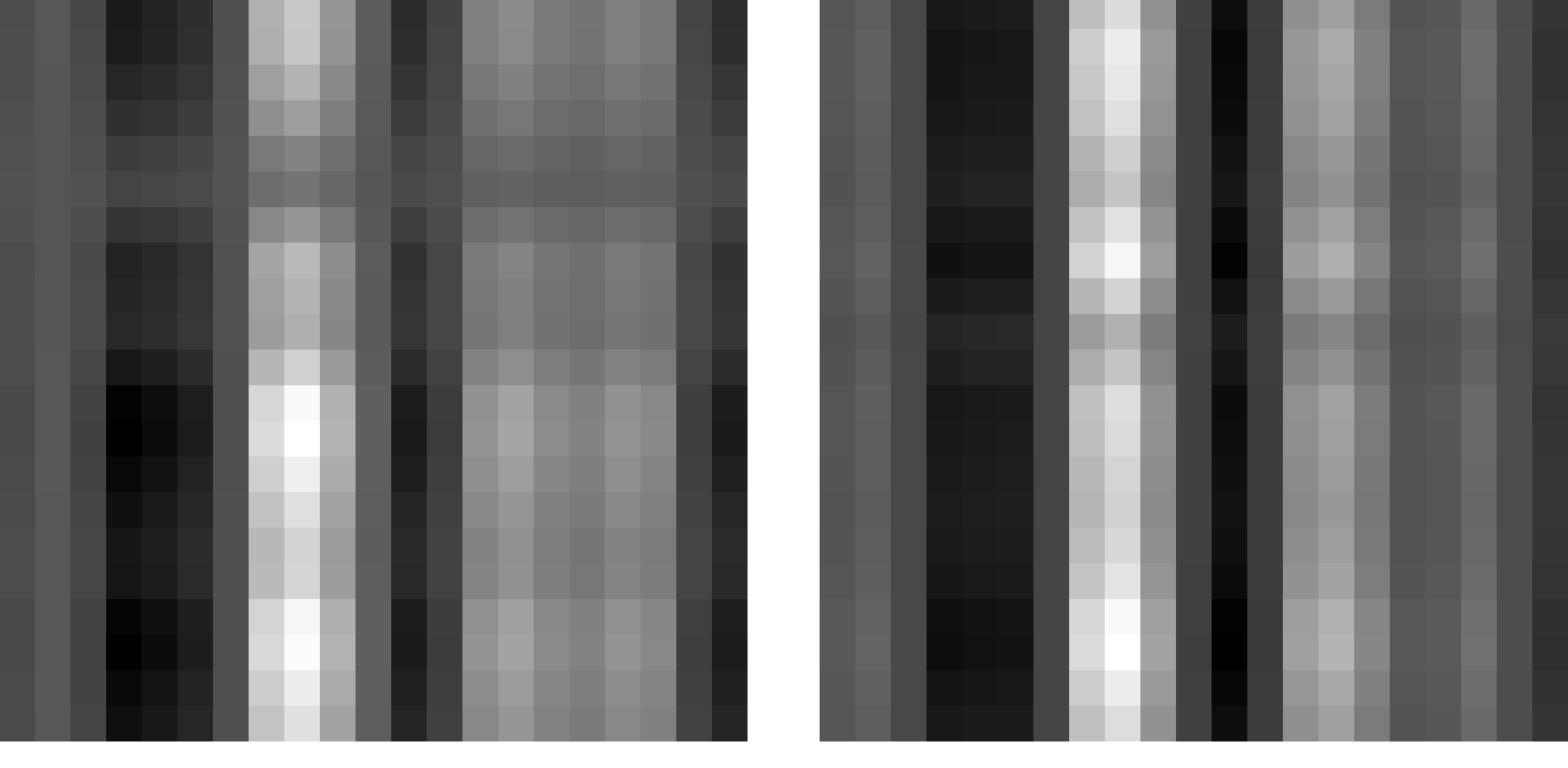} \hspace{-0.3em}
		  	\includegraphics[width=\separableFigyopWidth \textwidth, trim = 25 225 25 330, clip]{figures/separable/\nameData/separable_rank004_dim3.pdf} \hspace{-0.3em}

	}

	\vspace{-0.9em}
	\setcounter{subfigure}{1}
	\subfigure[Separable filters used for approximation]{
		  	\includegraphics[width=\separableFigyopWidth \textwidth, trim = 25 510 25 65, clip]{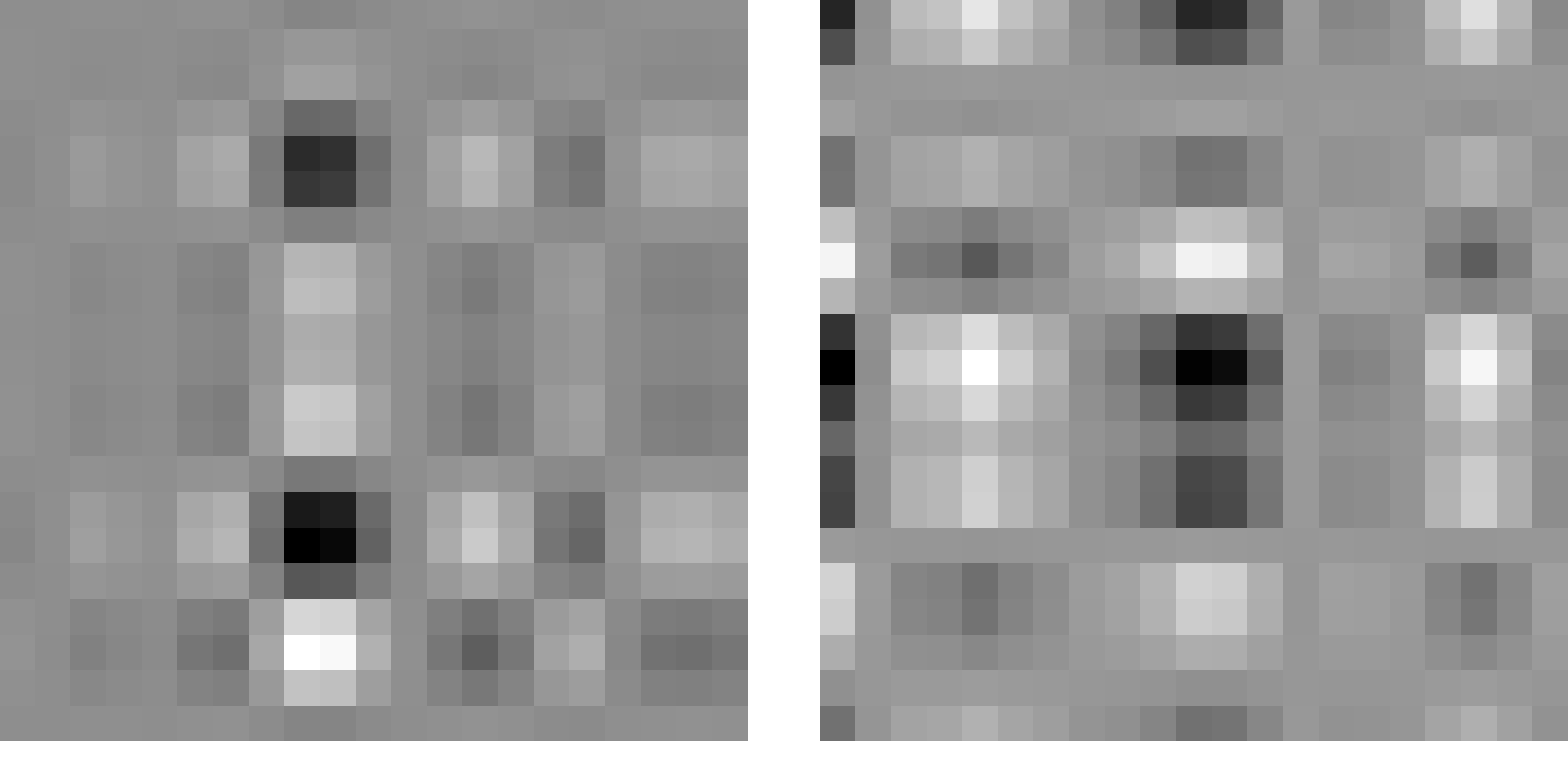} \hspace{-0.3em}
		  	\includegraphics[width=\separableFigyopWidth \textwidth, trim = 25 225 25 330, clip]{figures/separable/\nameData/separable_rank004_dim4.pdf} \hspace{-0.3em}

			\hspace{0.3em}
	
		  	\includegraphics[width=\separableFigyopWidth \textwidth, trim = 25 510 25 65, clip]{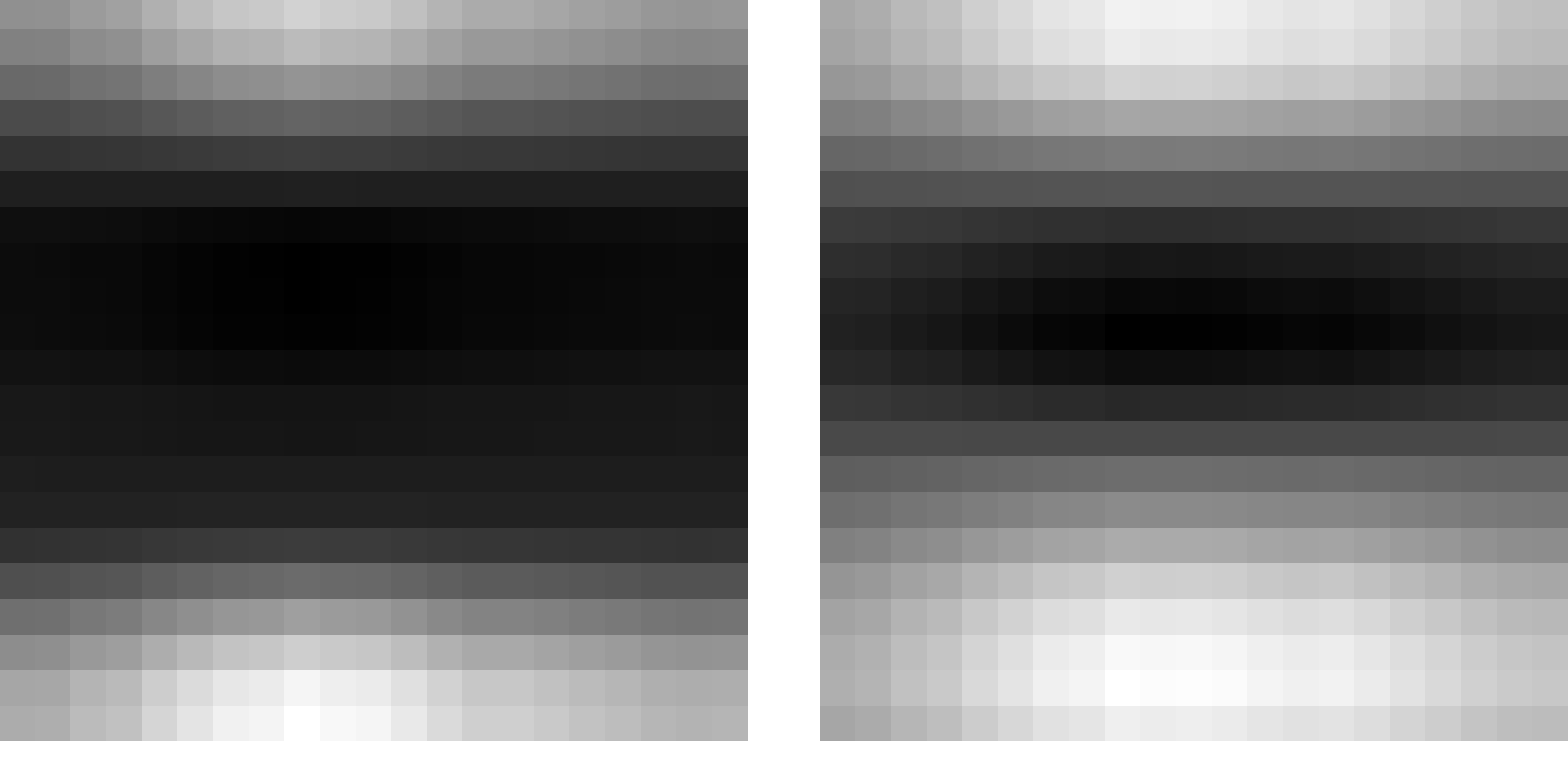} \hspace{-0.3em}
		  	\includegraphics[width=\separableFigyopWidth \textwidth, trim = 25 225 25 330, clip]{figures/separable/\nameData/separable_rank004_dim5.pdf} \hspace{-0.3em}

			\hspace{0.3em}

		  	\includegraphics[width=\separableFigyopWidth \textwidth, trim = 25 510 25 65, clip]{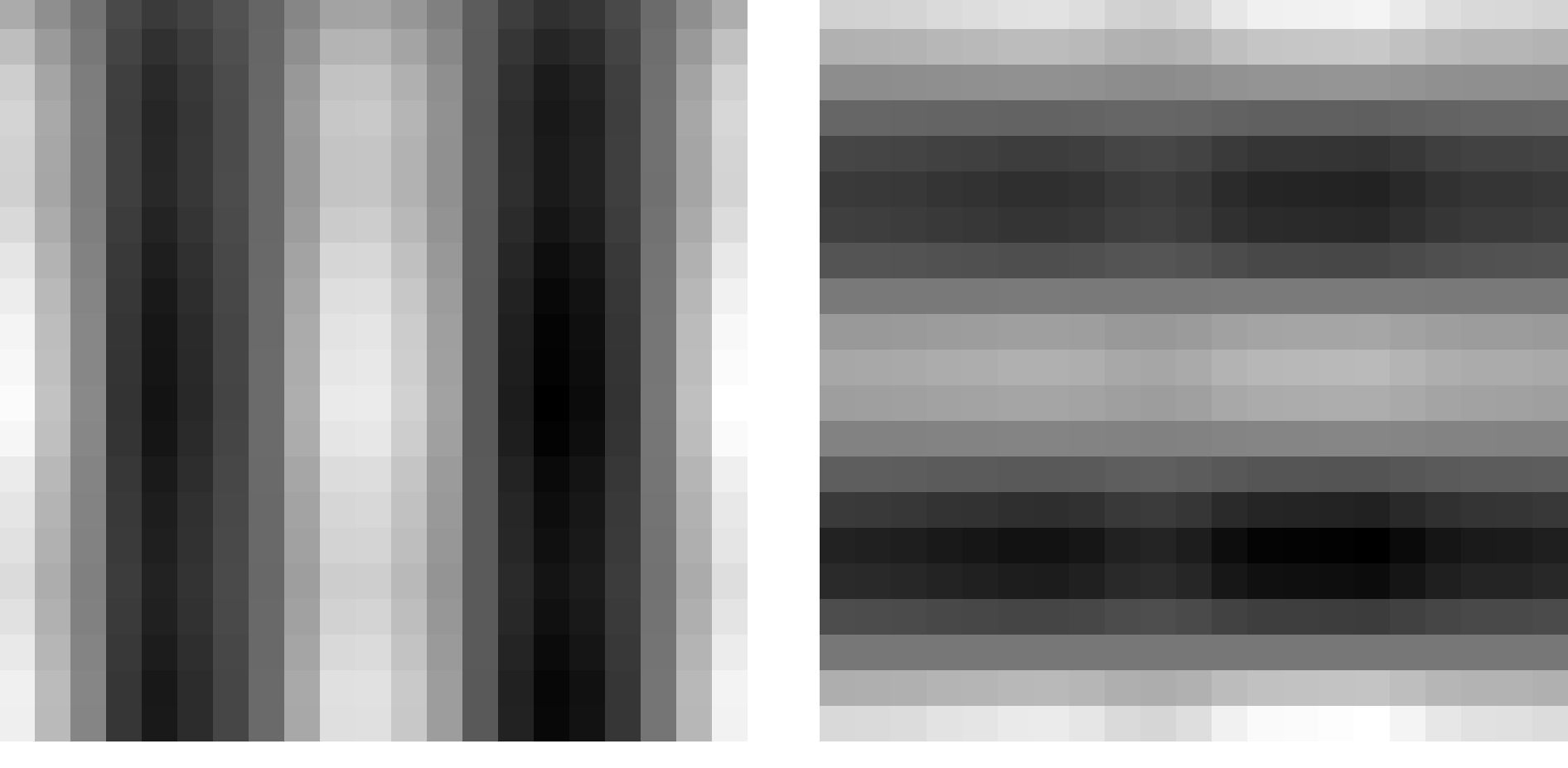} \hspace{-0.3em}
		  	\includegraphics[width=\separableFigyopWidth \textwidth, trim = 25 225 25 330, clip]{figures/separable/\nameData/separable_rank004_dim6.pdf} \hspace{-0.3em}

	\label{fig:separableApprox}	
	}

\caption{\subref{fig:separableOrig} The original 96   linear   filters  $\bw_{nm}$   learned   by   our  method   on
   the  \textit{StLouis} sequence. Each  row corresponds  to a  different image
   feature,  respectively  the horizontal  image  gradient,  the vertical  image
   gradient, the  magnitude of the gradient,  and the three color  components in
   the LUV  color space.  \subref{fig:separableApprox} The  24 separable filters
   learned    for    each    dimension   independently    using    the    method
   of~\cite{Sironi14b}. Each  original filter  can be  approximated as  a linear
   combination of the  separable filters, which can be convolved  with the input
   images very efficiently.}
\end{figure}

\paragraph{Optimization}

After dimension  reduction using  Principal Component Analysis~(PCA)  applied to
the training samples to decrease the  number of parameters to optimize, we solve
Eq.~\eqref{eq:problemFormulation} through a greedy procedure similar to gradient
boosting.  We start with an empty set of  hyperplanes $\bw_{n,m}$ and
we iteratively add new hyperplanes that minimize the objective function until we
reach  the desired  number (we  use  $N=4$ and  $M=4$ in  our experiments).   To
estimate   the   hyperplane  to   add,   we   apply   a  trust   region   Newton
method~\cite{Lin08}, as in the widely-used LibLinear library~\cite{Fan08}.

After initialization,  we randomly  go through  the hyperplanes  one by  one and
update them with the same Newton optimization method.  \fig{separableOrig} shows
the filters learned by our method  on the \textit{StLouis} sequence.  We perform
a  simple  cross-validation using  grid  search  in  log-scale to  estimate  the
meta-parameters $\gamma_c$, $\gamma_s$, and $\gamma_t$ on a validation set.


\paragraph{Approximation}

To further  speed up our  regressor, we  approximate the learned  linear filters
with  linear  combinations  of  separable  filters  using  the  method  proposed
in~\cite{Sironi14b}.   Convolutions  with  separable filters  are  significantly
faster  than convolutions  with  non-separable ones,  and  the approximation  is
typically   very  good.    \fig{separableApprox}  shows   an  example   of  such
approximated filters.

\section{Results}

\label{sec:results}

\begin{figure*}
\centering
\subfigure{
	\includegraphics[width=0.94\textwidth, height=82px, trim = 3 40 15 10, clip]{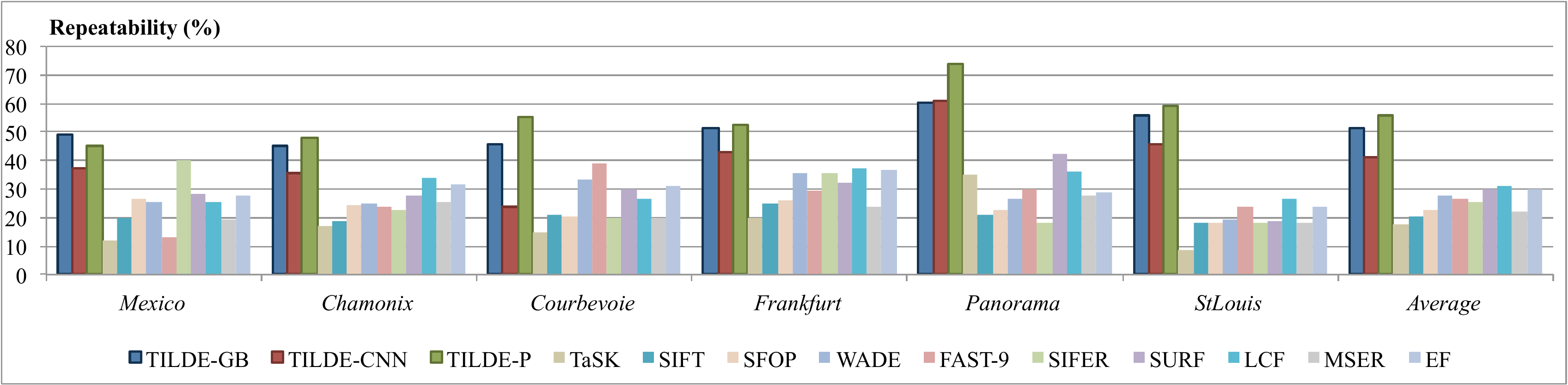}
}
	
\vspace*{-0.9em}
\subfigure{
	\includegraphics[width=0.94\textwidth, height=62px, trim =3 40 15 45, clip]{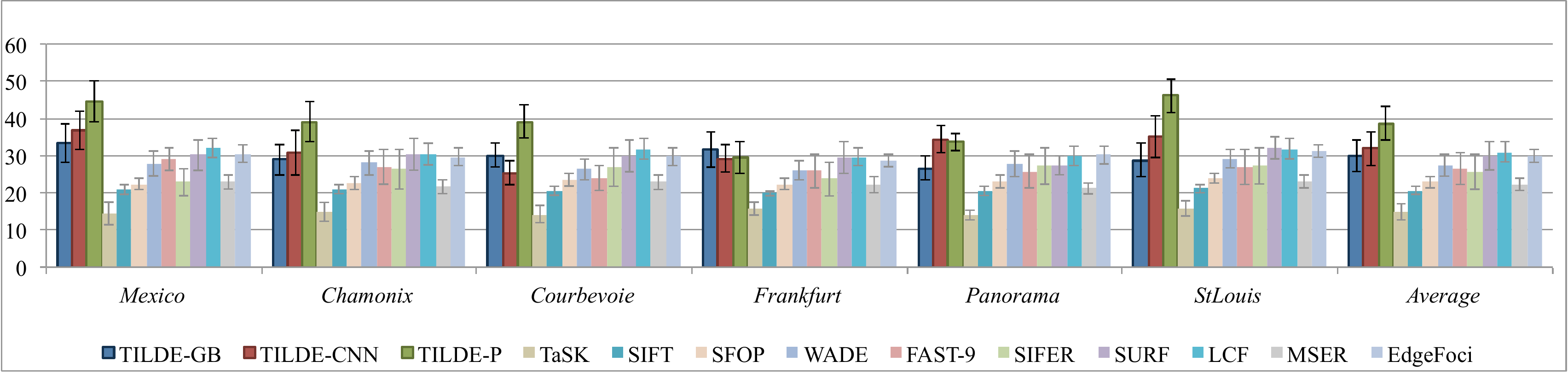}
}

\vspace*{-0.9em} 
\subfigure{
	\includegraphics[width=0.85\textwidth, trim =3 2 15 210, clip]{graphs/webcams/avg_repetability_cross.pdf}
}

\caption{Repeatability~(2\%) score on the {\it Webcam} dataset.  \textbf{Top:} average repeatability scores for each sequence trained on  the respective  sequences. \textbf{Bottom:}  average repeatability score when trained on one sequence (the name of the training sequence  is  given below each graph)  and tested  on all other sequences.  Although the  gap reduces  on the  bottom graph, our method significantly outperforms the  state-of-the-art in both  cases, which shows that our method can generalise to unseen scenes.}
\label{fig:rep}
\end{figure*}

In  this section  we  first describe  our experimental  setup  and present  both
quantitative  and   qualitative  results   on  our  \emph{Webcam}   dataset  and
the more standard \emph{Oxford} dataset.

\subsection{Experimental Setup}

We compare  our approach to  TaSK, SIFT, SFOP,  WADE, FAST-9, SIFER,  SURF, LCF,
MSER, and  EdgeFoci\footnote{See the  supplementary material  for implementation
  details.}.  
In the following, our full method will be denoted TILDE-P. TILDE-P24 denotes the
same method,  after approximation  of the piece-wise  linear regressor  using 24
separable filters.

To  evaluate  our regressor  itself,  we  also  compared  it against  two  other
regressors.   The  first  regressor,  denoted  TILDE-GB,  is  based  on  boosted
regression trees  and is an  adaptation of  the one used  in~\cite{Sironi14} for
centerline  detection to  keypoint  detection, with the same parameters used for
implementation as in the original work. The  second  regressor we  tried,
denoted  TILDE-CNN, is  a  Convolutional Neural  Network,  with an  architecture
similar  to   the  LeNet-5   network~\cite{LeCun98a}  but  with   an  additional
convolution layer and  a max-pooling layer.  The first, third,  and fifth layers
are convolutional layers;  the first layer has a resolution  of $28\times28$ and
filters  of size  $5\times5$,  the third  layer  has 10  features  maps of  size
$12\times12$ and filters of size $5\times5$, and the fifth layer 50 feature maps
of size  $4\times4$, and filters  of size  $3\times3$.  The second,  fourth, and
sixth layers are  max-pooling layers of size $2\times2$. The  seventh layer is a
layer of 500 neurons fully connected to the previous layer, which is followed by
the eighth  layer which  is a  fully-connected layer  with a  sigmoid activation
function, followed by the  final output layer.  For the output  layer we use the
$l_2$ regression  cost function.  


\subsection{Quantitative Results}

We  thoroughly  evaluated  the  performance  of  our  approach  using  the  same
repeatability measure as~\cite{Rosten10},  on our {\it Webcam}  dataset, and the
{\it Oxford} and {\it EF} datasets.   The repeatability is defined as the number
of   keypoints   consistently   detected   across  two   aligned   images.    As
in~\cite{Rosten10} we consider keypoints that are  less than 5 pixels apart when
projected to the same image as repeated.  However, the repeatability measure has
two  caveats: First,  a keypoint  close to  several projections  can be  counted
several times.  Moreover,  with a large enough number of  keypoints, even simple
random sampling can  achieve high repeatability as the density  of the keypoints
becomes high.

We therefore make  this measure more representative of the  performance with two
modifications: First, we allow a keypoint to be associated only with its nearest
neighbor,  in  other words,  a  keypoint  cannot be  used  more  than once  when
evaluating  repeatability.  Second,  we restrict  the number  of keypoints  to a
small given  number, so  that picking  the keypoints  at random  locations would
results   with    a   repeatability   score    of   only   2\%,    reported   as
\emph{Repeatability~(2\%)} in the experiments.

We also include  results using the standard repeatability  score, 1000 keypoints
per  image, and  a fixed  scale of  10 for  our methods,  which we  refer to  as
\emph{Oxford Stand.}  and \emph{EF Stand.}, for comparison with previous papers,
such  as~\cite{Mikolajczyk05, Zitnick11}.   \tbl{data}  shows a  summary of  the
quantitative results.

%
%
\begin{table}
  \centering
  \caption{Repeatability performance of our best regressors. The best results are in bold.  Our
    approach  provides  the highest  repeatability,  when  using our  piece-wise
    linear  regressor.  Note  that on {\it Oxford} and {\it EF} datasets  the performance  are
    slightly better when using smaller number of separable filters to
    approximate the original ones, probably  because the approximated filters
    tend to be smoother.}  \small
	\begin{tabularx}{0.47\textwidth}{l ccccc}
		\toprule
		& {\it Webcam} &  \multicolumn{2}{c}{{\it Oxford} }  &  \multicolumn{2}{c}{\it EF}  \\
	\#keypoints & {\it (2\%)} & {\it Stand.} & {\it (2\%)} & {\it Stand.} & {\it (2\%)} \\
		\midrule
	TILDE-GB 	& 33.3 		& 54.5 		& 32.8 		& 43.1 		& 16.2\\
	TILDE-CNN 	& 36.8 		& 51.8 		& 49.3 		& 43.2 		& 27.6 \\
	TILDE-P24 	& 40.7 		& \textbf{58.7} 	& \textbf{59.1}  & \textbf{46.3}	& \textbf{33.0} \\
	TILDE-P 		& \textbf{48.3} 	& 58.1 		& 55.9 		& {45.1} 		& 31.6 \\
\midrule
	FAST-9 		& 26.4 		& {53.8} 		& 47.9 		& 39.0 		& 28.0 \\
	SFOP 		& 22.9 		& 51.3		& 39.3 		& 42.2 		& 21.2 \\
	SIFER 		& 25.7 		& 45.1		& 40.1		& 27.4 		& 17.6 \\
	SIFT 		& 20.7 		& 46.5 		& 43.6 		& 32.2 		& 23.0 \\
	SURF 		& 29.9 		& 56.9 		& {57.6} 		& 43.6 		& 28.7 \\
	TaSK 		& 14.5 		& 25.7		& 15.7 		& 22.8 		& 10.0 \\
	WADE 		& 27.5 		& 44.3 		& 51.0 		& 25.6 		& 28.6 \\
	MSER 		& 22.3 		& 51.5 		& 35.9 		& 38.9 		& 23.9 \\
	LCF 			& {30.9} 		& 55.0 		& 40.1 		& 41.6	 	& 23.1 \\
	EdgeFoci 		& {30.0} 		& 54.9 		& 47.5 		& 46.2 		& 31.0 \\

		\bottomrule
	\end{tabularx}	

\label{tbl:data}%
\end{table}%

\subsubsection{Repeatability on our Webcam Dataset}


\fig{rep}  gives  the  repeatability   scores  for  our  \emph{Webcam}  dataset.
\fig{rep}-top shows the results of our  method when trained on each sequence and
tested on the same sequence, with the set of images divided into disjoint train,
validation, and test sets.  \fig{rep}-bottom shows the results when we apply our  detector trained  on  one  sequence to  all other unseen  sequences  from the   {\it Webcam}
dataset.   We significantly  outperform  state-of-the-art methods  when using  a
detector  trained specifically  to each  sequence.  Moreover,  while the  gap is
reduced when  we test  on un-seen  sequences, we  still outperform  all compared
methods by  a significant margin,  showing the generalization capability  of our
method.

\subsubsection{Repeatability on Oxford and EF Datasets}

\begin{figure*}
\begin{center}
  \includegraphics[width=0.94\textwidth, trim = 3 35 3 8, clip]{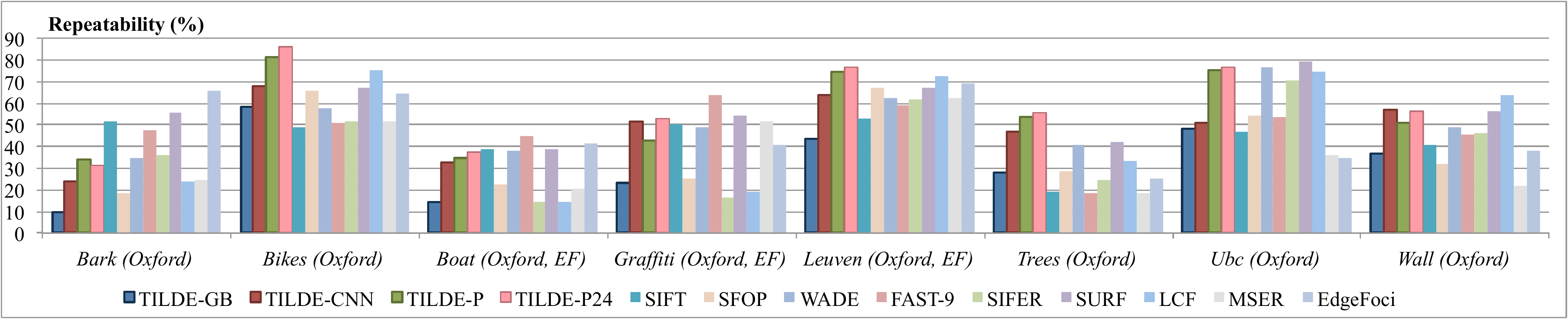}
\end{center}

\vspace{-0.6cm}
\begin{center}
  \includegraphics[width=0.94\textwidth, trim = 3 20 3 21, clip]{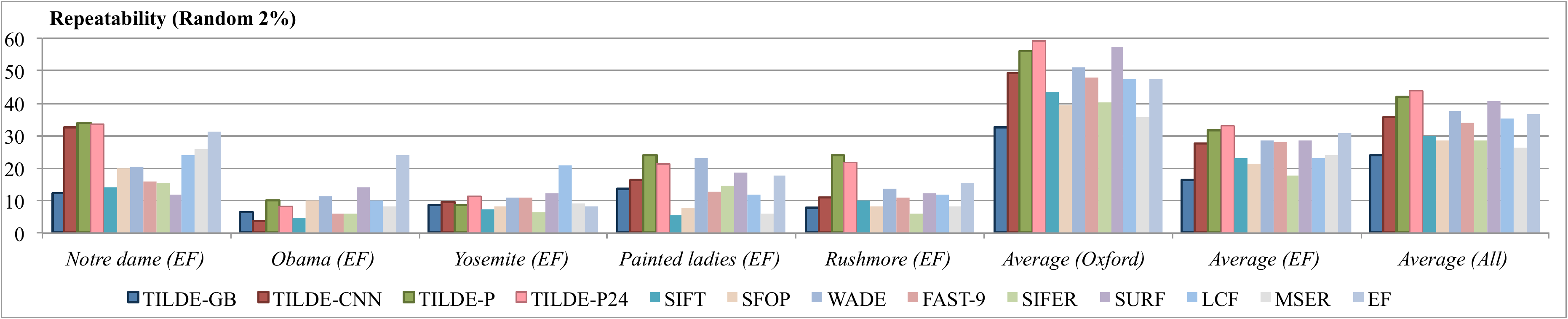}
\end{center}

\vspace{-0.9cm}
\begin{center}
  \includegraphics[width=0.94\textwidth, trim = 3 4 3 190, clip]{graphs/oxfords/cross.pdf}
\end{center}

\vspace{-0.2cm}
\caption{Repeatability~(2\%) score on the {\it Oxford} and {\it EF} datasets. Our methods
  are trained on the \textit{Chamonix} sequence from the {\it Webcam}
  dataset and tested on  {\it Oxford} and {\it EF} datasets.}
\label{fig:repeatabilityOxford}
\end{figure*}

In \fig{repeatabilityOxford}  we also evaluate  our method on \emph{Oxford} and  \emph{EF}
datasets.  \emph{Oxford}  dataset is  simpler in the  sense that it  does not  exhibit the
drastic changes of the  {\it Webcam} dataset but it is a  reference for the evaluation
of  keypoint detectors.  \emph{EF} dataset on the other hand exhibits drastic illumination changes and is very challenging. It is  therefore interesting  to evaluate our  approach on these datasets.

Instead  of learning  a new  keypoint  detector on  this dataset,  we apply  the
detector  learned  using the  \emph{Chamonix}  sequence  from the  \emph{Webcam}
dataset.   Our  method still  achieves  state-of-the-art  performance.  We  even
significantly   outperform  state-of-the-art   methods  in   the  case   of  the
\emph{Bikes}, \emph{Trees},  \emph{Leuven} and \emph{Rushmore} images,  which are outdoor scenes.
Note that we also obtain good results for \emph{Boat} which has large scale changes, although we currently do not consider scale in learning and detecting.
Repeatability score  shown here is  lower than
what  was  reported  in  previous  works~\cite{Mikolajczyk05,  Rosten10}  as  we
consider a smaller number of keypoints. As mentioned before, considering
a large number of keypoints artificially improves the repeatability score.

\begin{figure}
\centering
\includegraphics[width=0.44 \textwidth, trim = 3 11 6 6, clip]{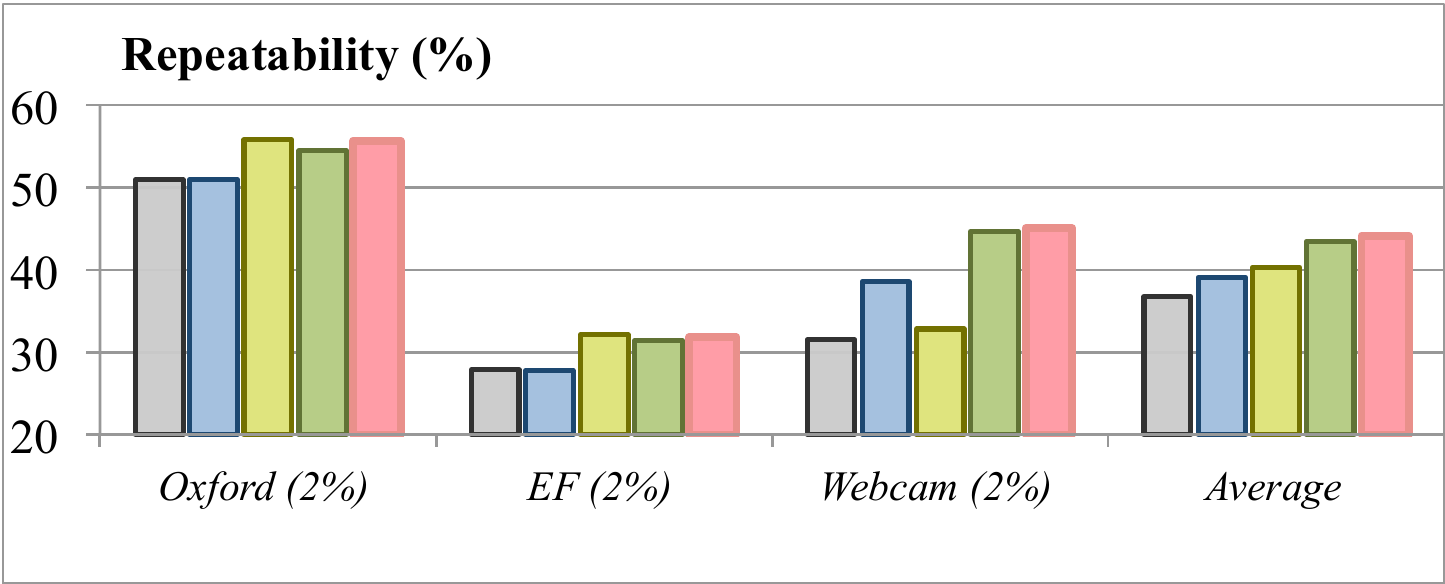}

\vspace{-0.3cm}
    \includegraphics[width=0.46\textwidth, trim = 3 11 6 150, clip]{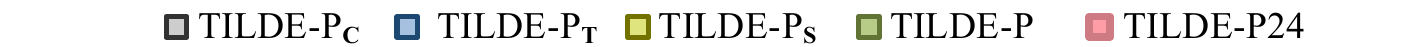}
\caption{\label{fig:eft}Effects of the three terms of  the
  objective function, and of the  approximation using separable filters. }
\end{figure}

\begin{figure}
\centering
 \includegraphics[width=0.42 \textwidth, trim = 19 42 5 4, clip]{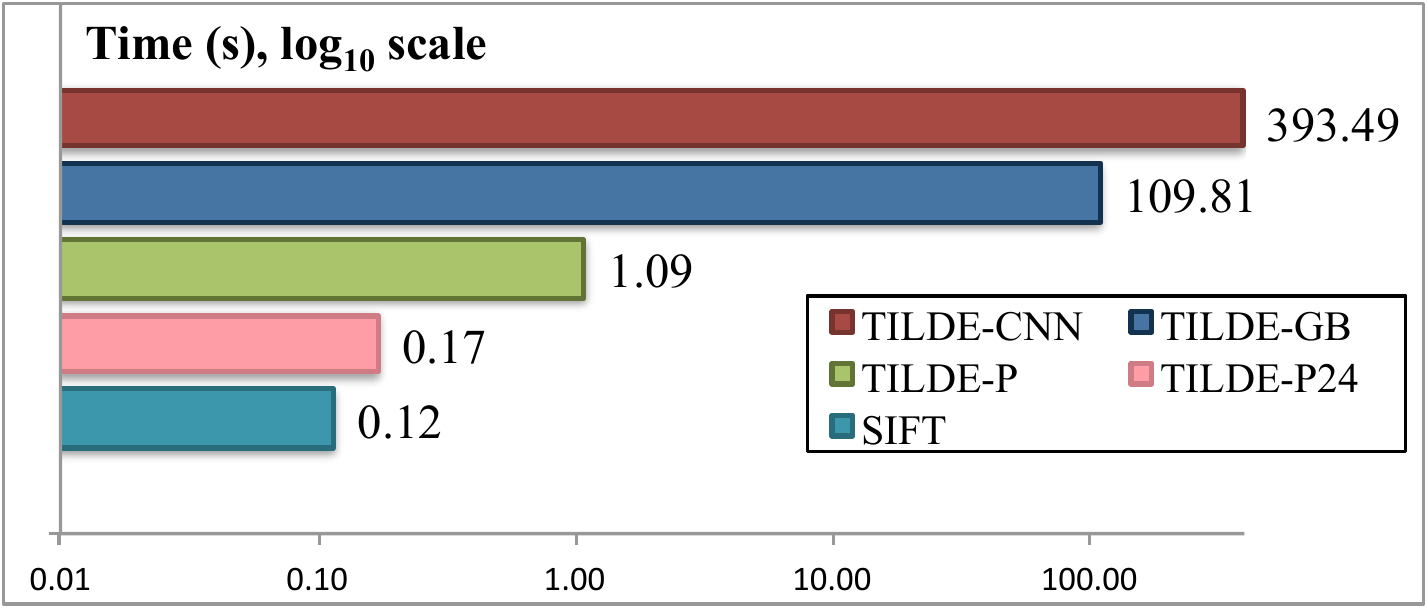}
\caption{\label{fig:time}Time comparison for the full pipeline of our various regressors compared with the SIFT detector. Evaluations were run on the same machine on an $640\times418$ image.}

\vspace{-0.8em}

\end{figure}

\subsection{Qualitative Results}

\def \QualitativeFigHeight {4.2cm}
\begin{figure*}
	\centering
	\subfigure[Original images]{
		  	\includegraphics[height=\QualitativeFigHeight, trim = 25 20 1210 24, clip]{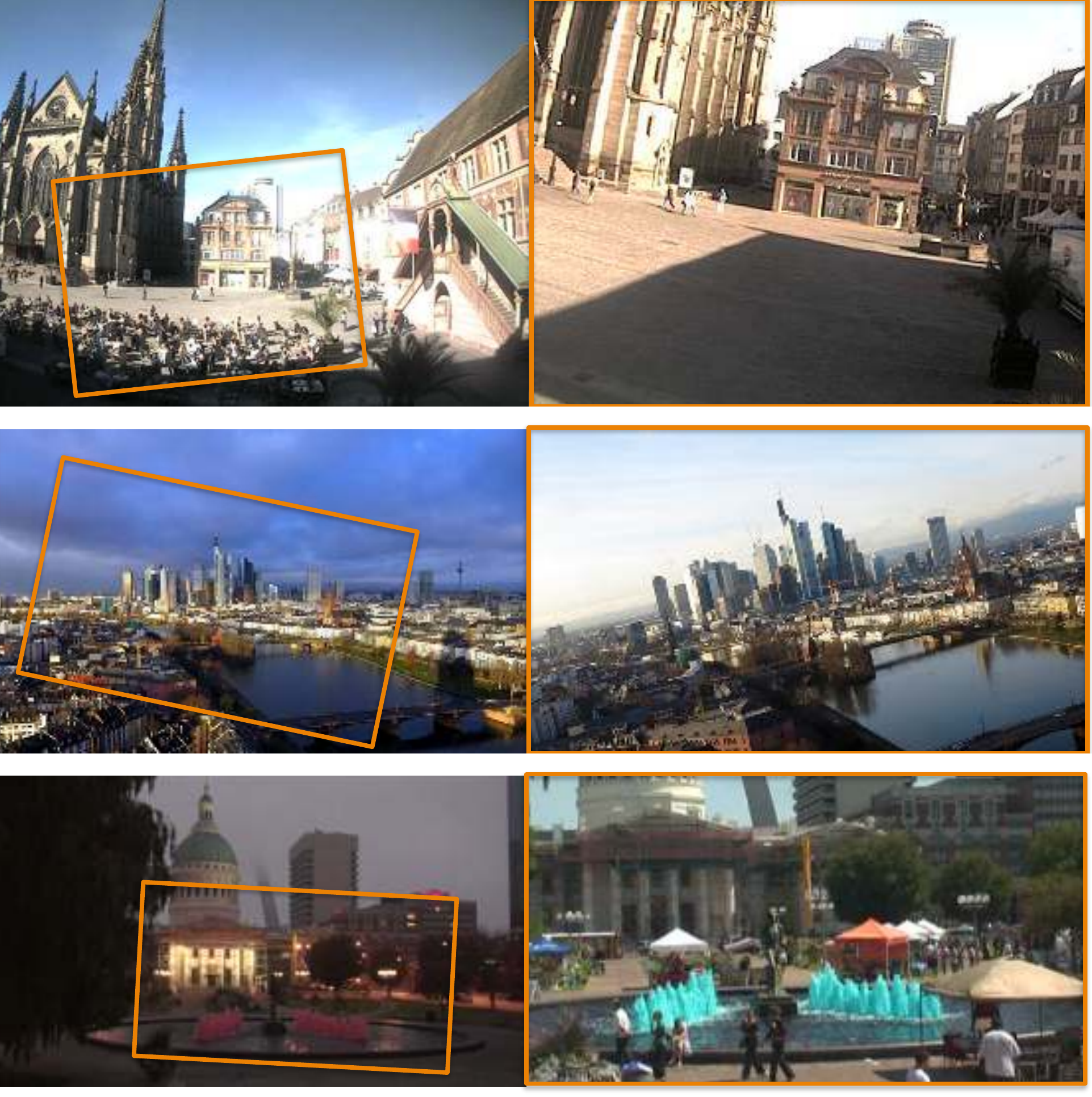}
		  	\label{fig:qualitativeOrig}
	}
	\subfigure[SIFT]{
		  	\includegraphics[height=\QualitativeFigHeight, trim = 600 20 925 24, clip]{figures/qualitative/qualitative_img.pdf}
		  	\label{fig:qualitativeSIFT}
	}
	\hspace*{-0.6em}
	\subfigure[SURF]{
		  	\includegraphics[height=\QualitativeFigHeight, trim = 900 20 625 24, clip]{figures/qualitative/qualitative_img.pdf}
		  	\label{fig:qualitativeSURF}
	}
	\hspace*{-0.6em}
	\subfigure[FAST-9]{
		  	\includegraphics[height=\QualitativeFigHeight, trim = 1200 20 325 24, clip]{figures/qualitative/qualitative_img.pdf}
		  	\label{fig:qualitativeFAST}
	}
	\hspace*{-0.6em}
	\subfigure[Our keypoints]{
		  	\includegraphics[height=\QualitativeFigHeight, trim = 1500 20 25 24, clip]{figures/qualitative/qualitative_img.pdf}
		  	\label{fig:qualitativeTILDE}
	}

   	\caption{Qualitative results on several images from different
          sequences.  From top to bottom: \textit{Courbevoie},
          \textit{Frankfurt}, and
          \textit{StLouis}. \subref{fig:qualitativeOrig} Pairs of
          images to be matched, with ground truth transformation,
          transformations obtained with \subref{fig:qualitativeSIFT}
          the SIFT detector, \subref{fig:qualitativeSURF} the SURF
          detector, \subref{fig:qualitativeFAST} the FAST-9 detector,
          and \subref{fig:qualitativeTILDE} our TILDE
          detector.}  \label{fig:qualitative}
\end{figure*}

We  also give  in  \fig{qualitative} some  qualitative results  on  the task  of
matching challenging pairs of images  captured at different days under different
weather  conditions.  Our  matching  pipeline  is as  follow:  we first  extract
keypoints in both images using the different methods we want to compare, compute
the keypoints  descriptors, and  compute the homography  between the  two images
using RANSAC.   Since the goal of  this comparison is to  evaluate keypoints not
descriptors, we  use the  SIFT descriptor  for all methods.   Note that  we also
tried                                 using                                other
descriptors~\cite{Bay08,Rublee11,Calonder10b,Alahi12,Leutenegger11}  but due  to
the drastic  difference between the  matched images, only SIFT  descriptors with
ground  truth orientation  and scale  worked.  We  compare our  method with  the
SIFT~\cite{Lowe04},    SURF~\cite{Bay08},   and    \mbox{FAST-9}~\cite{Rosten10}
detectors, using the same number of keypoints~(300) for all methods.  Our method
allows to retrieve the correct  transformations between the images even under
such drastic changes of the scene appearance.

\subsection{Effects of the Three Loss Terms}
\label{sec:expThreeTerm}




\fig{eft} gives the results of the evaluation of the influence of each loss term
of Eq.~\eqref{eq:problemFormulation} by evaluating the performance of our detector without
each term.  We will refer to our  method when using only the classification loss
as  $\text{TILDE-P}_{\text{C}}$, when  using  both classification  loss and  the
temporal  regularization  as  $\text{TILDE-P}_{\text{T}}$, and  when  using  the
classification loss and the shape regularization as $\text{TILDE-P}_{\text{S}}$.
We achieve  the best performance when  all three terms are  used together.  Note
that the  shape regularization  enhances the  repeatability on  \emph{Oxford} and  \emph{EF}, two
completely unseen datasets, whereas  the temporal  regularization helps  when we
test on images which are similar to the training set.

\subsection{Computation Times}

\fig{time} gives the computation time of SIFT and each variant of our method. TILDE-P24 is not very far from SIFT. Note that our
method  is highly  parallelizable, while  our current  implementation does  not
benefit from  any parallelization. We therefore  believe that our method  can be
significantly sped up with a better implementation.

\section{Conclusion}

We have introduced a learning scheme  to detect keypoints reliably under drastic
changes of weather and lighting conditions.  We proposed an effective method for
generating the training  set to learn regressors.  We  learned three regressors,
which  among them,  the  piece-wise  linear regressor  showed  best result.   We
evaluated our  regressors on  our new outdoor  keypoint benchmark  dataset.  Our
regressors  significant  outperforms the  current  state-of-the-art  on our  new
benchmark dataset and also achieve state-of-the-art performances on {\it Oxford}
and {\it EF} datasets, demonstrating their generalisation capability.

An  interesting future  research  direction is  to extend  our  method to  scale
space. For example, the strategy  applied in~\cite{Leutenegger11} to FAST can be
directly applied to our method.

%

\section*{Acknowledgement}
This work was supported by the EU FP7 project MAGELLAN under the grant number ICT-FP7-611526 and in part by the EU project EDUSAFE.

{\small
\bibliographystyle{ieee}
\bibliography{string,vision,learning}
}


\end{document}